\documentclass[acmtog,nonacm,table]{acmart}

\AtBeginDocument{%
  }

\definecolor{intermimicyellow}{RGB}{235,190,45}

\usepackage{xspace}
\newcommand{\eg}{\textit{e.g.}\@\xspace}

\usepackage{array}
\usepackage{enumitem}
\usepackage{subcaption}
\usepackage{bbding}
\usepackage{fontawesome5}
\usepackage{adjustbox}
\usepackage[normalem]{ulem}

\definecolor{cornflowerblue}{rgb}{0.39, 0.58, 0.93}

\begin{document}

\title{LYRIC: Language-Driven Physics-Based Character Control for Contact-Rich Whole-Body Object Interaction}

\author{Zeyu Han}
\affiliation{
  \institution{Northeastern University}
  \city{Boston}
  \country{USA}
}

\author{Zichong Meng}
\affiliation{
  \institution{Northeastern University}
  \city{Boston}
  \country{USA}
}

\author{Julian Tanke}
\affiliation{
  \institution{Sony Corporate Technology Center America Inc}
  \city{New York}
  \country{USA}
}

\author{Minami Matsumoto}
\affiliation{
  \institution{Sony Interactive Entertainment Inc}
  \city{Minato}
  \state{Tokyo}
  \country{Japan}
}

\author{Sergey Bashkirov}
\affiliation{
  \institution{Sony Interactive Entertainment Inc}
  \city{San Mateo}
  \country{USA}
}

\author{Yingruo Fan}
\affiliation{
  \institution{Sony Corporation of America}
  \city{San Mateo}
  \country{USA}
}

\author{Selim Engin}
\affiliation{
  \institution{Sony Corporation of America}
  \city{San Mateo}
  \country{USA}
}

\author{Dongseok Shim}
\affiliation{
  \institution{Sony Group Corporation}
  \city{Minato}
  \state{Tokyo}
  \country{Japan}
}

\author{Takashi Shibuya}
\affiliation{
  \institution{Sony Group Corporation}
  \city{Minato}
  \state{Tokyo}
  \country{Japan}
}

\author{Yuki Mitsufuji}
\affiliation{
  \institution{Sony Corporate Technology Center America Inc}
  \city{New York}
  \country{USA}
}

\author{Huaizu Jiang}
\affiliation{
  \institution{Northeastern University}
  \city{Boston}
  \country{USA}
}

\renewcommand{\shortauthors}{Han et al.}

\authorsaddresses{}

\begin{abstract}
We present \textbf{LYRIC}, a generative flow-matching controller for \textbf{L}anguage-driven ph\textbf{Y}sics-based contact-\textbf{R}ich \textbf{I}nteraction \textbf{C}ontrol, that enables simulated characters to perform contact-rich whole-body object interactions from a free-form language instruction and a sparse terminal object goal.
To obtain reliable expert trajectories from imperfect motion-capture references, a single tracking policy is trained using \emph{geometry-conditioned interaction rewards} and relaxed reference tracking near hand--object contact.
To guide interaction progress without prescribing a full-body kinematic reference, we factorize the controller into a task-level planner that predicts short-horizon object and humanoid-root trajectories, and an action generator that resolves whole-body motion and contacts in closed loop.
After behavior cloning, we freeze the planner and post-tune the action generator on policy using the planner's predictions as stable supervision for intermediate task progression.
In a controlled OMOMO evaluation, our tracker achieves 64.3\% success compared with 53.2\% for an InterMimic reimplementation,
while a unified policy achieves 76.5\% on the full OMOMO dataset.
On the held-out split, LYRIC achieves 90.3\% task success, compared with 74.2\% for the strongest matched kinematic-planner baseline, with better semantic alignment and motion quality.
Without retraining, the controller also supports test-time object-waypoint guidance.
Qualitative results further demonstrate robust, natural contact-rich interactions and zero-shot transfer to novel object shapes.
The webpage is available at \href{https://neu-vi.github.io/LYRIC/}{https://neu-vi.github.io/LYRIC/}.
\end{abstract}


\begin{teaserfigure}
  \centering
  \includegraphics[width=\textwidth]{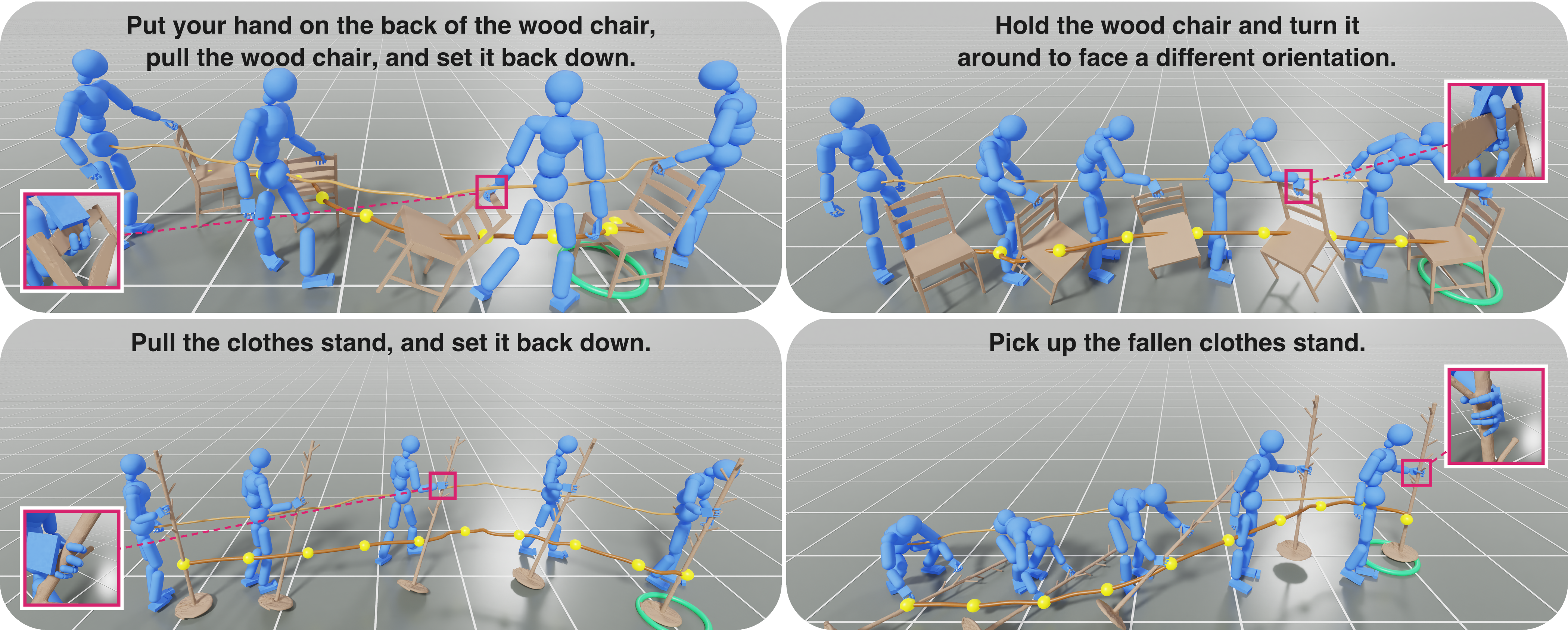}
  \caption{Given only a text prompt and a sparse terminal object goal, 
  our method produces physically simulated
  whole-body object interactions: the character approaches, grasps, carries, and
  places diverse objects with realistic hand contact. 
  }
  \label{fig:teaser}
\end{teaserfigure}

\maketitle
\section{Introduction}
 
Controlling physically simulated characters to perform diverse, natural behaviors is a central goal in character animation, and recent advances in reinforcement learning have brought remarkable progress to this pursuit.
Natural language provides an intuitive interface for directing character behaviors, allowing users to specify both what the character should accomplish and how the motion should be performed.
A key remaining frontier is contact-rich object interaction: enabling simulated characters to pick up, carry, and manipulate objects with coordinated whole-body motion and dexterous hands. 
We aim to control characters to perform the desired interaction from only a free-form language instruction and a sparse terminal object goal as shown in Fig.~\ref{fig:teaser}.
 
Recent work on physics-based character control has extended whole-body policies to object interaction~\cite{wang2023physhoi,xu2025intermimic,xu2026interprior,tessler2025maskedmanipulator}, which are directed through dense motion references or structured 3D goals (\eg, desired object poses).
While these inputs specify useful spatial guidance, free-form language provides a complementary and intuitive means for expressing interaction semantics and manner of execution.
Due to the rapid progress of text-conditioned kinematic generators~\cite{xu2023interdiff,li2023object,peng2025hoi,li2024controllable,peng2026surface}, a natural idea to support language instructions is placing them upstream of a tracking controller, as in recent work of character control for locomotion~\cite{wu2025human,tevet2024closd,lin2025simgenhoi}.
For contact-rich interactions, however, a generated human--object trajectory may be kinematically plausible yet difficult to realize under physical dynamics and contact constraints, particularly because the object is unactuated.
Closed-loop replanning~\cite{tevet2024closd} and mutual fine-tuning~\cite{lin2025simgenhoi} improve the coupling between generation and execution while retaining a dense full-body trajectory as the interface to physical control.
The quality of the generated reference can therefore become a bottleneck, since errors in hand--object coordination or contact timing can make the prescribed interaction difficult to realize even with a capable tracking controller.

Generative control policies offer a promising alternative by modeling a distribution over motor actions conditioned directly on the language instruction, task goal, and current simulated state.
They can adapt body motion and contacts as object interactions unfold without following a separately generated full-body trajectory at deployment.
Such policies are commonly initialized by imitating expert controllers via behavior cloning or policy distillation~\cite{truong2024pdp,huang2025diffuse,wu2025uniphys,zhang2026script,tessler2025maskedmanipulator,xu2026interprior}.
Subsequent refinement commonly uses DAgger-style imitation or on-policy reinforcement learning~\cite{tessler2025maskedmanipulator,ren2024diffusion,zhang2025reinflow,chen2025pirl,zhang2026script,xu2026interprior} to improve robustness beyond the state distribution covered by offline expert demonstrations.
However, applying this paradigm to contact-rich whole-body object interaction presents two challenges.
\emph{First}, imitating experts requires high-quality rollouts.
For whole-body object interaction, expert controllers are typically trained to track motion-capture kinematic references, whose finger poses may be unreliable near hand--object contact, thereby compromising the resulting expert rollouts.
\emph{Second}, on-policy post-tuning needs a task signal for how the instructed interaction should unfold.
Yet neither the language instruction nor the terminal object goal provides sufficient supervision for intermediate task progression.

In this paper, we present \textbf{LYRIC}, a generative controller for \textbf{L}anguage-driven ph\textbf{Y}sics-based contact-\textbf{R}ich \textbf{I}nteraction \textbf{C}ontrol that enables contact-rich whole-body object interaction from a free-form language instruction and a sparse terminal object goal.
To address the first challenge, we propose \emph{geometry-conditioned interaction rewards}, which adapts grasp supervision to local object thickness, to train a single tracking policy from imperfect motion-capture references.
Near demonstrated contact, the tracker relaxes reference tracking for the hands and supporting upper-body joints, allowing it to deviate from unreliable captured hand poses while preserving the remaining body and object motion.
Together, these mechanisms produce physically executed expert rollouts with reliable hand interaction.
To address the second challenge, we factorize the controller into two models: a task-level trajectory planner and a whole-body action generator, both parameterized as conditional flow matching Transformers~\cite{lipman2022flow,peebles2023scalable}.
The planner predicts short-horizon object and humanoid-root trajectories that express the intended object motion and accompanying locomotion without prescribing joint poses or contacts.
The action generator realizes these task-level plans as whole-body actions through closed-loop execution.
We train both flow models by behavior cloning on successful tracking rollouts, then freeze the planner and post-tune the action generator on policy.
During post-tuning, the frozen planner's predictions provide stable, instruction-conditioned object--root supervision beyond the terminal goal, while simulation-based objectives evaluate physical execution.
At deployment, the planner and action generator operate without the tracking policy or a prescribed dense full-body motion.

Our experiments demonstrate reliable contact-rich motion tracking and language-driven character control. 
In a controlled subject-specific comparison on three OMOMO~\cite{li2023object} subjects, our tracking formulation achieves $64.3\%$ success, compared with $53.2\%$ for an InterMimic reimplementation~\cite{xu2025intermimic}, demonstrating the benefit of our interaction rewards and contact-dependent tracking relaxation.
Our unified tracker achieves a $76.5\%$ tracking success rate on the full OMOMO using a single policy, providing abundant, high-quality rollouts for behavior cloning.
Our language-driven generative controller achieves a $90.3\%$ task success rate on the held-out split, compared with $74.2\%$ for the strongest kinematic-planner baseline with a fine-tuned tracker, while also improving text--motion alignment and overall motion quality.
We further demonstrate control through user-specified object waypoints and zero-shot transfer to novel object shapes.

Our contributions are summarized as follows:
\begin{itemize}[itemsep=0pt,topsep=0pt,leftmargin=18pt]
    \item We propose \emph{geometry-conditioned interaction rewards} that adapt grasps to local object geometry. Along with relaxed reference tracking, a single tracking policy is developed to convert imperfect motion-capture references into physically executed expert trajectories with reliable hand interaction.
 
    \item We introduce a flow-matching controller for language-driven contact-rich whole-body object interaction, comprising a task-level trajectory planner and a whole-body action generator. The planner's object--root trajectories guide closed-loop execution and provide intermediate supervision for on-policy post-tuning while leaving joint motion and contacts to the action generator.
 
    \item Controlled evaluations show higher tracking success than InterMimic~\cite{xu2025intermimic} and better task success, instruction adherence, and motion quality than text-conditioned kinematic-planner baselines. We further demonstrate test-time object-waypoint guidance and qualitative zero-shot transfer to novel within-category shapes.
\end{itemize}

\section{Related Work}

\paragraph{Physics-Based Character and Object Interaction Control.}
Physics-based character control learns motor policies under simulated dynamics, ranging from reference-motion imitation to reusable motion priors and high-level language- or goal-conditioned control~\citep{peng2018deepmimic,peng2021amp,peng2022ase,yao2022controlvae,luo2023universal,juravsky2022padl,juravsky2024superpadl,tessler2024maskedmimic}.
Recent work extends these controllers to object interaction through reference tracking, structured interaction specifications, or geometry-grounded control representations~\citep{wang2023physhoi,xu2025intermimic,tessler2025maskedmanipulator,xu2026interprior,lin2026lessmimic,liang2026interreal}.
Language-conditioned planner--tracker systems use generated kinematic character--object trajectories to guide physical execution, but remain constrained by full-body references whose character and object trajectories may not be jointly realizable under contact dynamics~\citep{wu2025human,lin2025simgenhoi}.

Constructing the tracking expert also requires reliable contact supervision.
Existing trackers commonly use reference-derived contacts, body-object distances, object-tracking objectives, or generic grasp priors~\citep{wang2023physhoi,xu2025intermimic,tessler2025maskedmanipulator,xu2026interprior}.
Recent methods introduce explicit contact commands, wrist-guided finger learning, or geometry-aware interaction representations~\citep{li2026contactmimic,yu2026wristmimic,lin2026lessmimic}.
Our geometry-conditioned interaction rewards instead make the desired hand-object relationship depend directly on local geometry, encouraging enclosure around thin structures and support beneath flat surfaces to produce more natural and stable expert trajectories.

\paragraph{Kinematic Human-Object Interaction Generation from Language.}
Text-conditioned human motion generation has progressed rapidly with diffusion models, autoregressive motion tokens, and motion--language pretraining~\citep{zhang2024motiondiffuse,tevet2022human,zhang2023generating,jiang2023motiongpt,meng2025absolute,meng2025rethinking}.
Recent methods extend language-conditioned motion generation to object interaction by generating coordinated human and object trajectories from language and other interaction cues, using object geometry, contact prediction, affordance reasoning, or physics-informed guidance to improve interaction quality~\citep{xu2023interdiff,li2023object,diller2024cg,li2024controllable,peng2025hoi,xu2024interdreamer,yang2024f,ron2025hoidini,wu2025hoi,peng2026surface}.
FlowHOI further applies flow matching to generate language-grounded hand poses, object poses, and contact states for downstream robot retargeting~\citep{zeng2026flowhoi}.
Nevertheless, these methods primarily produce kinematic interaction sequences rather than closed-loop actions for a physically simulated humanoid.
Our method instead maps language and the current simulated state directly to whole-body actions, allowing the controller to revise its motion and contact strategy during execution.

\paragraph{Generative Policies for Physics-Based Control.}
Diffusion and flow policies provide expressive action distributions for multimodal control~\citep{janner2022planning,ajay2022conditional,chi2025diffusion,lipman2022flow}.
In character animation, PDP directly generates actions, CLoSD couples a text-conditioned motion planner with a physics tracker, Diffuse-CLoC predicts state-action look-ahead distributions, and UniPhys integrates planning and control in a unified diffusion framework~\citep{truong2024pdp,tevet2024closd,huang2025diffuse,wu2025uniphys}.
These methods primarily address object-free motion or retain a generated full-body reference.
SCRIPT trains a text-conditioned diffusion action policy through flow matching and improves it with online RL, but focuses on object-free humanoid motion~\citep{zhang2026script}.
Conversely, InterPrior combines generative character control with RL for object manipulation, but is conditioned on structured interaction goals rather than language~\citep{xu2026interprior}.
Our method brings these directions together in a text-conditioned flow-matching controller that coordinates whole-body motion and dexterous object manipulation in physics simulation.
Rather than tracking a dense full-body reference, the planner in our model predicts only short-horizon object and humanoid-root trajectories to guide the closed-loop action generator, leaving detailed body motion and contact strategies to the controller.

\paragraph{RL for Diffusion and Flow Policies.}
Recent work has developed RL methods for post-training diffusion and flow policies.
DDPO formulates diffusion denoising as a multi-step decision process, while DPPO adapts policy-gradient fine-tuning to continuous robot control~\citep{black2023training,ren2024diffusion}.
Flow-based methods make online optimization tractable through stochastic flow formulations, likelihood estimation, or flow-matching surrogate objectives~\citep{zhang2025reinflow,mcallister2026flow,yi2026flow,yang2026policyflow,chen2025pirl}.
Related VLA systems combine language-conditioned flow action models with RL alignment for dexterous robot manipulation~\citep{apanasevich2026green}.
Our focus is character control, where the policy must coordinate whole-body motion and dexterous manipulation of a dynamically simulated object.
We adopt the \emph{Flow-SDE} formulation from $\pi_{\mathrm{RL}}$, which converts deterministic flow sampling into a stochastic process suitable for PPO-based online refinement~\citep{chen2025pirl}.
We derive dense task rewards from the planner's predicted object and humanoid-root trajectories and use on-policy post-tuning to improve recovery from contact-induced state deviations during execution.
To our knowledge, this is the first online RL post-training of a text-conditioned flow-matching controller for contact-rich character control, jointly coordinating locomotion and dexterous manipulation of an unactuated object in physics simulation.

\begin{figure*}[!t]
    \centering
    \includegraphics[width=0.95\textwidth]{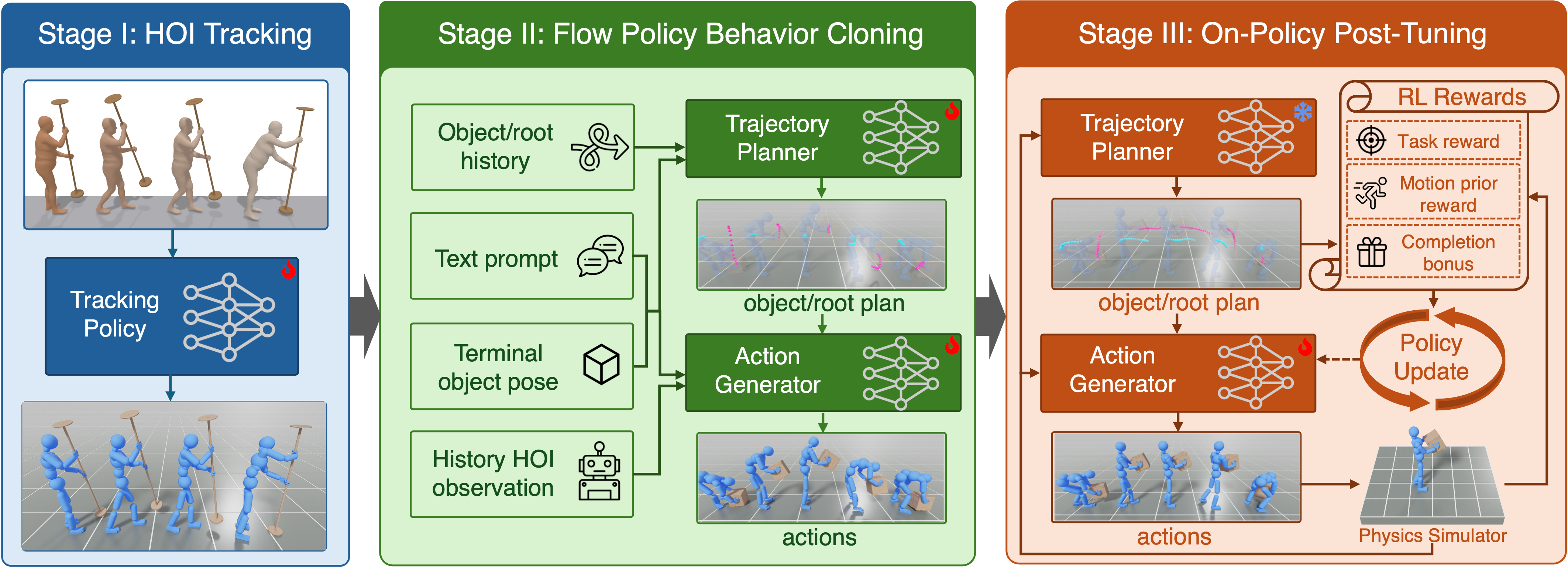}
    \caption{\textbf{Training pipeline for LYRIC.}
    We first train a physics-based HOI tracking expert from kinematic demonstrations.
    We then roll out the expert to collect physically grounded state--action trajectories and train the trajectory planner and action generator offline by behavior cloning.
    Finally, we freeze the planner and post-tune the action generator on policy in simulation using planner-derived plan-following rewards together with interaction, style, and terminal-placement objectives.
    We use {\color{cornflowerblue}{\faSnowflake}} to mark frozen components and {\color{red}{\faFire}} to mark trainable components.
    } 
    \label{fig:method}
\end{figure*}

\section{Methodology}
\label{sec:method}

We first formulate the language-driven object interaction control task and provide an overview of our three-stage pipeline (Sec.~\ref{sec:formulation}).
Each stage is then described in detail: geometry-conditioned interaction tracking (Sec.~\ref{sec:tracking}), factorized text-conditioned control (Sec.~\ref{sec:flow}), and on-policy post-tuning (Sec.~\ref{sec:flow_rl}).

\subsection{Overview}
\label{sec:formulation}

\paragraph{Task formulation.}
We consider a physics-based humanoid character, represented as SMPL-X~\cite{smpl}, interacting with a single unactuated 
object.
The humanoid has 51 actuated 3-DoF joints, of which 30 belong to the two hands.

The interaction unfolds as a discrete-time control loop.
At control step $t$, we define the state of the humanoid--object system as
\begin{equation}
\begin{aligned}
    \mathbf{s}_t &= [\mathbf{s}^{\mathrm H}_t,\mathbf{s}^{\mathrm O}_t],\\
    \mathbf{s}^{\mathrm H}_t
    &= [h_t,\mathbf{p}^{\mathrm H}_t,\mathbf{R}^{\mathrm H}_t,
        \mathbf{v}^{\mathrm H}_t,\boldsymbol{\omega}^{\mathrm H}_t],\\
    \mathbf{s}^{\mathrm O}_t
    &= [\mathbf{p}^{\mathrm O}_t,\mathbf{R}^{\mathrm O}_t,
        \mathbf{v}^{\mathrm O}_t,\boldsymbol{\omega}^{\mathrm O}_t],
\end{aligned}
\label{eq:state}
\end{equation}
where $h_t$ is the humanoid root height, and $\mathbf{p}$, $\mathbf{R}$, $\mathbf{v}$, and $\boldsymbol{\omega}$ denote positions, 6D rotations, and linear and angular velocities.
The superscripts $\mathrm{H}$ and $\mathrm{O}$ indicate whether a quantity belongs to the humanoid or to the object.
Contact-rich control further depends on how the two bodies touch, so from the state and the object geometry we derive interaction features $\boldsymbol{\phi}_t$, including humanoid--object contact states, signed distances and surface directions, and local object-shape descriptors, whose exact composition is specified in the appendix.
The controller's observation is $\mathbf{o}_t=[\mathbf{s}_t;\boldsymbol{\phi}_t]$, and its action $\mathbf{a}_t\in\mathbb{R}^{153}$ specifies joint-position targets tracked by joint-level PD controllers; the simulator advances the state through its dynamics $\mathcal{F}$ as $\mathbf{s}_{t+1}=\mathcal{F}(\mathbf{s}_t,\mathbf{a}_t)$.
Unless stated otherwise, we express positions, vectors, and orientations in the humanoid-root-centric heading frame, aligned with the root's yaw, and measure root and object heights relative to the ground.

Given an initial state $\mathbf{s}_0$, a natural-language instruction $y$ that specifies the interaction semantics and manner, and a desired terminal object pose $g$ that constrains the object only at the end of the interaction, the task is to find a controller $\pi$ that generates actions
\begin{equation}
    \mathbf{a}_t
    \sim
    \pi\!\left(
        \,\cdot\mid
        \mathbf{o}_{\leq t},
        \mathbf{a}_{<t},
        y,
        g
    \right),
    \label{eq:task}
\end{equation}
where $\mathbf{o}_{\leq t}=(\mathbf{o}_0,\ldots,\mathbf{o}_t)$ and $\mathbf{a}_{<t}=(\mathbf{a}_0,\ldots,\mathbf{a}_{t-1})$ collect the past observations and actions.
The induced closed-loop rollout should realize the instructed interaction and bring the object to $g$.
Because the object is unactuated, all task-directed object motion must be produced through contact with the humanoid.

\paragraph{Proposed method.}
Fig.~\ref{fig:method} summarizes the three-stage training pipeline of LYRIC.
Motion-capture datasets of human-object interaction record natural, contact-rich whole-body strategies and pair each sequence with a language description.
Such demonstrations are kinematic though, specifying the intended human and object motion but not the motor commands or contact forces that realize it.
Stage~I therefore trains a tracking policy with geometry-conditioned interaction rewards (Sec.~\ref{sec:tracking}) and rolls it out under observation perturbations to broaden coverage around each demonstration, converting the demonstrations into a dataset of physically executed state--action trajectories.

Stage~II trains the controller from this dataset via behavior cloning (Sec.~\ref{sec:flow}).
We design the controller to also support subsequent on-policy post-tuning, which improves robustness to simulated states outside the distribution covered by offline expert rollouts.
Such post-tuning, however, requires task supervision for how the interaction should unfold.
Rewarding only the terminal object placement does not capture whether the object is manipulated as specified by the language instruction, while the instruction itself does not provide per-step spatial targets.
We therefore factorize the controller into a planner that predicts short-horizon object and humanoid-root trajectories and a whole-body action generator that realizes these plans.
The predicted trajectories condition action generation and also provide per-step task targets for post-tuning.
The same instruction admits many valid interactions, so we model both the planner and the action generator as distributions over their outputs with flow matching~\citep{lipman2022flow} and sample one coherent interaction at test time.

Stage~III realizes this post-tuning by freezing the planner and post-tuning the action generator on policy in simulation, using the predicted object--root trajectories to construct dense task rewards alongside interaction, motion-style, and terminal-placement objectives (Sec.~\ref{sec:flow_rl}).
These planner-derived targets guide intermediate interaction progress, while interaction, motion-style, and terminal-placement objectives evaluate the realized execution in simulation.
At deployment, the frozen planner and the post-tuned action generator together realize $\pi$, without any tracking policy, kinematic reference, or externally provided object trajectory or contact sequence.
A user-specified object trajectory can optionally replace the planner's plan for spatial control (Sec.~\ref{sec:exp_waypoint}).

\subsection{Geometry-Conditioned Interaction Tracking}
\label{sec:tracking}

\paragraph{Tracking policy and reward.}
As shown in Stage I of Fig.~\ref{fig:method}, we use a single tracking policy to convert each demonstration into physically executed state--action trajectories.
Interaction tracking conditions the policy on a kinematic demonstration, which serves as the reference, and rewards it for reproducing the demonstrated humanoid and object motion under the simulator dynamics $\mathcal{F}$.
Following InterMimic~\citep{xu2025intermimic}, the input at control step $t$ comprises the observation $\mathbf{o}_t$ and residuals from the current state $\mathbf{s}_t$ to the reference states $\hat{\mathbf{s}}$ (the hat marks reference quantities) at two future frames.
We further add the last action $\mathbf{a}_{t-1}$, which tells the policy what it most recently commanded, and a one-hot object identity, which lets a single policy specialize its behavior across objects.
Whereas InterMimic trains one teacher policy per subject and distills them into a unified student, our formulation enables a single policy trained directly over all subjects and objects.

\begin{figure}[!t]
    \centering
    \includegraphics[width=0.5\textwidth]{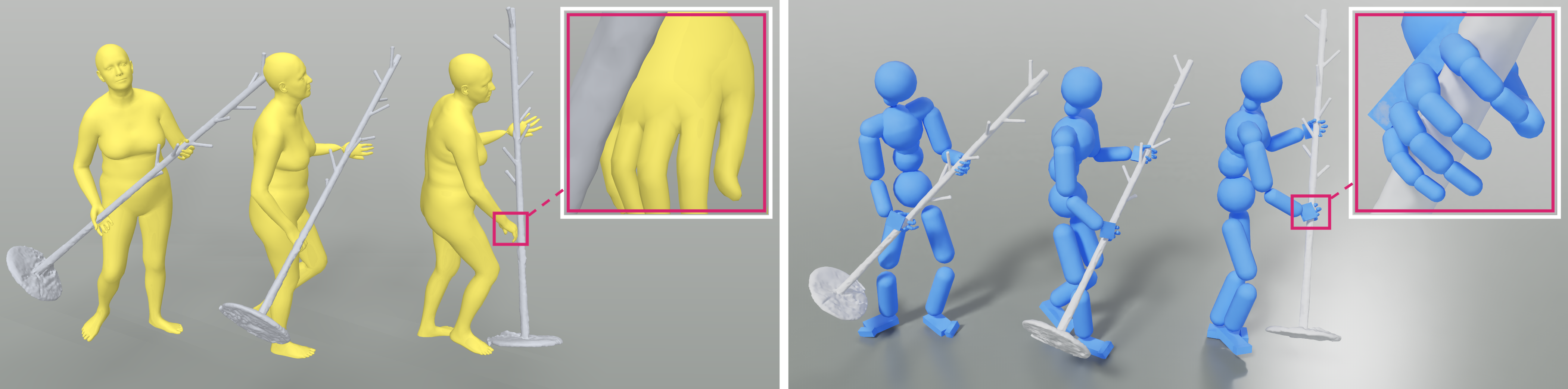}
    \caption{\textbf{Correcting imperfect hand--object interactions.}
    Given imperfect kinematic hand poses (\textcolor{yellow!80!black}{left}), our geometry-conditioned interaction objectives allow the tracker to form natural and stable hand motion while largely preserving the demonstrated whole-body motion (\textcolor{blue!70}{right}).
    }
    \label{fig:grasp_rw_motivation}
\end{figure}

The per-step tracking reward combines three groups of nonnegative costs and peaks when all costs vanish:
\begin{equation}
    r^{\mathrm{track}}_t
    =
    \exp\!\left[
        -\left(
            E^{\mathrm{ref}}_t
            +E^{\mathrm{grasp}}_t
            +E^{\mathrm{reg}}_t
        \right)
    \right].
    \label{eq:tracking_reward}
\end{equation}
Here, $E^{\mathrm{ref}}_t$ is the reference-tracking cost from~\cite{xu2025intermimic}, which penalizes deviation of the simulated humanoid, the object, and their contacts from the reference at the current step.
Matching the reference hand poses and contacts alone, however, does not ensure a grasp that can hold the object due to the noisy motion-captured data.
We therefore introduce $E^{\mathrm{grasp}}_t$, a geometry-conditioned grasp cost that scores the arrangement of the hand's contacts against the local object geometry. %
$E^{\mathrm{reg}}_t$ collects the regularization terms.
Figure~\ref{fig:grasp_rw_motivation} illustrates how our interaction objectives refine imperfect reference hand poses while largely preserving the demonstrated whole-body motion.
Below, we describe how $E^{\mathrm{ref}}_t$ is relaxed during contact and then introduce $E^{\mathrm{grasp}}_t$ and $E^{\mathrm{reg}}_t$.
Additional details are given in the appendix.

\paragraph{Gated reference tracking.}
Motion-captured interaction data often contain imperfect finger poses, especially around object contact, although the captured hand motion remains a useful reference away from contact.
We therefore retain hand-pose tracking during free motion and relax it during contact, resorting to the geometry-conditioned grasp cost to refine the hand pose toward a more stable grasp.
Because the refined hand pose may deviate from its reference, the supporting arm may also need to adjust while still following the demonstrated arm motion overall.
We keep the reference-tracking terms for the remaining body parts and the object unchanged to preserve the demonstrated whole-body interaction.
Specifically, we introduce a per-hand reference-contact gate $\eta\in[0,1]$ (0 means no contact at all), computed from reference contacts over a short temporal window.
The gate affects three groups of body parts differently:
\begin{itemize}[leftmargin=16pt,itemsep=1pt,topsep=2pt]
    \item \textbf{Fingers and wrist.} Their reference-tracking terms in $E^{\mathrm{ref}}_t$ are weighted by $1-\eta$, while the corresponding per-hand grasp cost introduced below is weighted by $\eta$. At $\eta=0$, these reference terms retain their full weights and the grasp cost has zero weight. At $\eta=1$, these reference terms have zero weight and the grasp cost has its full weight.
    \item \textbf{Elbow, shoulder, and thorax.} Their reference-tracking terms are scaled by $1-\eta(1-w)$, where $w\in(0,1)$ is a fixed part-specific fraction that increases from the elbow through the shoulder to the thorax. Reference tracking therefore relaxes progressively less farther up the kinematic chain.
    \item \textbf{All other parts and the object.} Their reference-tracking terms retain their original weights for all values of $\eta$.
\end{itemize}

\paragraph{Geometry-conditioned grasp cost.}
As the gate relaxes hand-pose tracking when the reference indicates contact, the grasp cost must judge the resulting grasp on its own merits.
Our key insight is that the object's local shape constrains what a \emph{functional} grasp can look like.
We can thus score the geometric prerequisites of a grasp from the shape and the contacts the hand makes on it.
A thin structure, such as a handle or pole, affords enclosure, where the fingers wrap around it and press from opposing sides.
In contrast, a broad surface affords support, where the hand spreads its contacts to stabilize the object.
Our cost therefore scores each hand by two complementary criteria that encode necessary geometric conditions for a stable grasp, one for enclosure and one for support.

\begin{figure}[!t]
    \centering
    \includegraphics[width=0.45\textwidth]{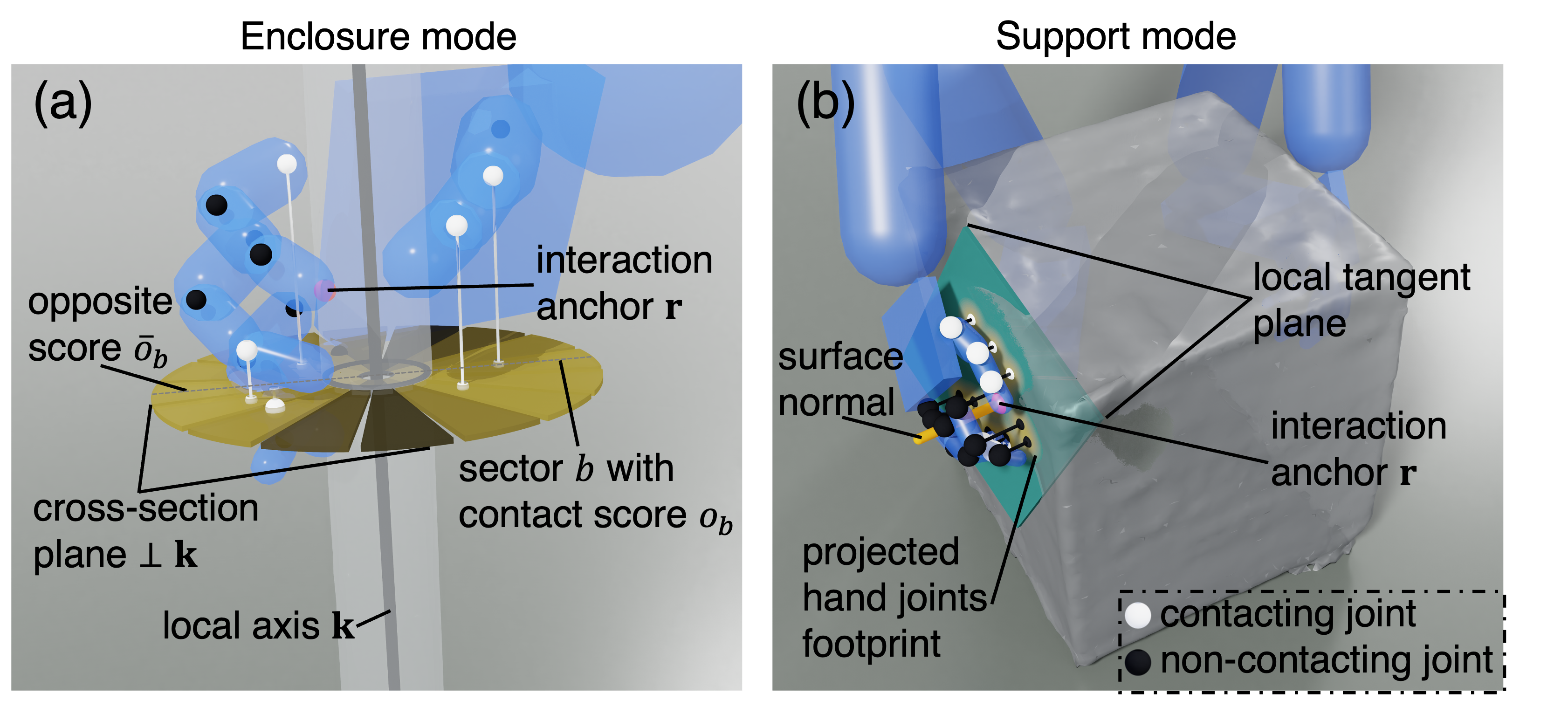}
    \caption{\textbf{Geometry-conditioned grasp objectives.} Enclosure rewards opposing contacts on thin structures, while support rewards distributed contact on broad surfaces.
    }
    \label{fig:grasp}
\end{figure}

To simplify notation, all quantities below refer to one hand at a control step, omitting the hand and time indices.
The per-hand grasp cost is defined as
\begin{equation}
    e^{\mathrm{grasp}}
    =
    \alpha\left(1-Q^{\mathrm{enc}}\right)
    +
    (1-\alpha)\left(1-Q^{\mathrm{sup}}\right),
    \label{eq:geometry_blend}
\end{equation}
where $Q^{\mathrm{enc}},Q^{\mathrm{sup}}\in[0,1]$ measure enclosure and support quality as defined below.
$\alpha\in[0,1]$ weights the two modes and is computed 
around an interaction anchor $\mathbf{r}$, the object-surface point nearest to the centroid of the reference hand, as shown in Fig.~\ref{fig:grasp}.
We set $\alpha$ by the local object thickness around $\mathbf{r}$, read from a precomputed thickness field akin to the shape diameter function~\citep{shapira2008consistent}.
The complete term $E^{\mathrm{grasp}}_t$ weights each hand's $e^{\mathrm{grasp}}$ by its gate $\eta$ and sums over both hands.

\textbf{Enclosure mode.}
Fig.~\ref{fig:grasp}(a) illustrates $Q^{\mathrm{enc}}$ for a thin local structure, where a stable grasp requires squeezing it between the thumb and the fingers, engaging several fingers, and placing contacts on opposing sides of its cross section.
We estimate the local long axis $\mathbf{k}$ of the structure as the direction of least variation among the nearby surface normals.
A cross-sectional plane orthogonal to $\mathbf{k}$, centered at the midpoint of the local thickness profile, is then constructed.
Partitioning the plane into $B$ uniform angular sectors, we project the contacting finger joints onto it and score each sector by the contact it receives.
The score $o_b\in[0,1]$ rises from $0$ toward $1$ as contacts accumulate in the sector's direction, with soft assignment to neighboring sectors and saturating aggregation of repeated contacts.
The opposite score $\bar{o}_b\in[0,1]$ reads the same measure on the far side of the cross section, around the direction opposite to sector $b$ (the dashed line in Fig.~\ref{fig:grasp}(a)).
The enclosure quality is defined as
\begin{equation}
    Q^{\mathrm{enc}}
    =
    G^{\mathrm{tf}}
    G^{\mathrm{part}}
    \frac{
        \sum_b o_b\bar{o}_b
    }{
        \sum_b o_b
    },
    \label{eq:enclosure_score}
\end{equation}
which multiplies three terms, one per requirement above.
The thumb--finger gate $G^{\mathrm{tf}}$ increases with the contact of the thumb and of the other fingers and vanishes when either is absent, since the squeeze needs the thumb opposing the fingers.
The participation gate $G^{\mathrm{part}}$ grows with the number of distinct non-thumb fingers in contact and reaches one when all four participate, since more fingers make the hold more secure.
The fraction is high only when contacts have counterparts on the opposite side, which distinguishes enclosing the structure from merely touching it.

\textbf{Support mode.}
Fig.~\ref{fig:grasp}(b) illustrates $Q^{\mathrm{sup}}$ for a broad surface, which the hand cannot enclose, so a stable support requires physical contact with the surface and a footprint spread over it.
We estimate the local surface normal around $\mathbf{r}$ and take the tangent plane orthogonal to it. 
We project all hand joints onto this plane, producing the two-dimensional footprint as shown in Fig.~\ref{fig:grasp}(b), whose covariance $\boldsymbol{\Sigma}_{\mathrm{tan}}$ gives an area-like measure of its extent through $\sqrt{\det\boldsymbol{\Sigma}_{\mathrm{tan}}}$.
The support quality is defined as
\begin{equation}
    Q^{\mathrm{sup}}
    =
    G^{\mathrm{contact}}\,
    \psi\!\left(
        \sqrt{
            \det\!\left(
                \boldsymbol{\Sigma}_{\mathrm{tan}}
            \right)
        }
    \right),
    \label{eq:support_score}
\end{equation}
which multiplies two terms, one per requirement above.
The contact-coverage gate $G^{\mathrm{contact}}$ increases with the number of hand joints physically contacting the object and vanishes without contact, which prevents a widely spread but hovering hand from receiving a high support score.
The spread term $\psi$ normalizes the footprint extent to $[0,1]$, reaching $0$ for a collapsed footprint and $1$ for a fully spread hand, which prevents a compact cluster of contacts from being treated as effective support.

\paragraph{Regularization terms.}
The enclosure and support modes evaluate the spatial arrangement of hand--object contacts, but they do not by themselves cover the entire interaction process.
Therefore, in addition to InterMimic's energy costs on joint and object accelerations and on contact forces, which discourage jerky motion and excessive force, we add three hand-interaction regularization terms to $E^{\mathrm{reg}}_t$, listed below.
\begin{itemize}[leftmargin=16pt,itemsep=1pt,topsep=2pt]
    \item \textbf{Palm alignment.} We encourage the palm to face the local object surface, guiding contact formation and correcting wrist-orientation errors common in motion-captured interaction data.
    
    \item \textbf{Contact anchor.} We penalize the distance between the simulated hand-contact region and the corresponding object-local contact region indicated by the kinematic reference. This preserves where the demonstrated interaction occurs while allowing the geometry-conditioned objective to refine how the hand grasps the object.
    
    \item \textbf{Hold stability.} Once contact is established, we penalize sliding of contacting finger joints in the object frame and the loss of previously established finger contacts. New contacts are not penalized, so the term specifically discourages grasp slip and contact breaking.
\end{itemize}

\paragraph{Tracking-policy training and rollout collection.}
We train the tracking policy with PPO~\citep{schulman2017proximal} to maximize the expected return under Eq.~\eqref{eq:tracking_reward}.
During training, we randomly perturb the policy's observation $\mathbf{o}_t$, preparing it for the observation noise used during rollout collection.
We terminate an episode when the humanoid, the object, or their distance field deviates from the reference beyond a threshold, when the humanoid falls, or when a body--object contact present in the reference is lost in simulation.

After training, we generate multiple rollouts for each demonstration by executing the policy under independently resampled observation noise.
Resampling the noise produces varied state--action trajectories around each demonstration.
Only rollouts that reach the end of the demonstration without early terminations are retained.
Each retained rollout records the states $\mathbf{s}_t$ and executed actions $\mathbf{a}_t$ at every control step and is paired with its terminal object pose $g$ and the language instruction $y$ of the demonstration it tracks, forming the training data for Stage~II.

\subsection{Factorized Text-Conditioned Controller}
\label{sec:flow}

\paragraph{Controller factorization.} 
As shown in Stage II of Fig.~\ref{fig:method}, we factorize the controller $\pi$ of Eq.~\eqref{eq:task} into the trajectory planner $\pi_{\mathrm P}$ and the whole-body action generator $\pi_{\mathrm A}$.
Both are parameterized as conditional flow-matching transformers~\citep{lipman2022flow}, with $L_{\mathrm P}$ and $L_{\mathrm A}$ blocks for the planner and action generator, respectively.
The planner predicts only the object and humanoid-root trajectories, without joint poses of the humanoid or contact labels, and leaves the body motion and contacts to the action generator.
These trajectories condition action generation and are later reused as task supervision for on-policy post-tuning (Sec.~\ref{sec:flow_rl}).
It can also serve as the interface for spatial control, where a user-specified object or root trajectory can replace it at deployment (Sec.~\ref{sec:exp_waypoint}) to guide the resulting motion along a desired path.

\begin{figure}[!t]
    \centering
    \includegraphics[width=0.45\textwidth]{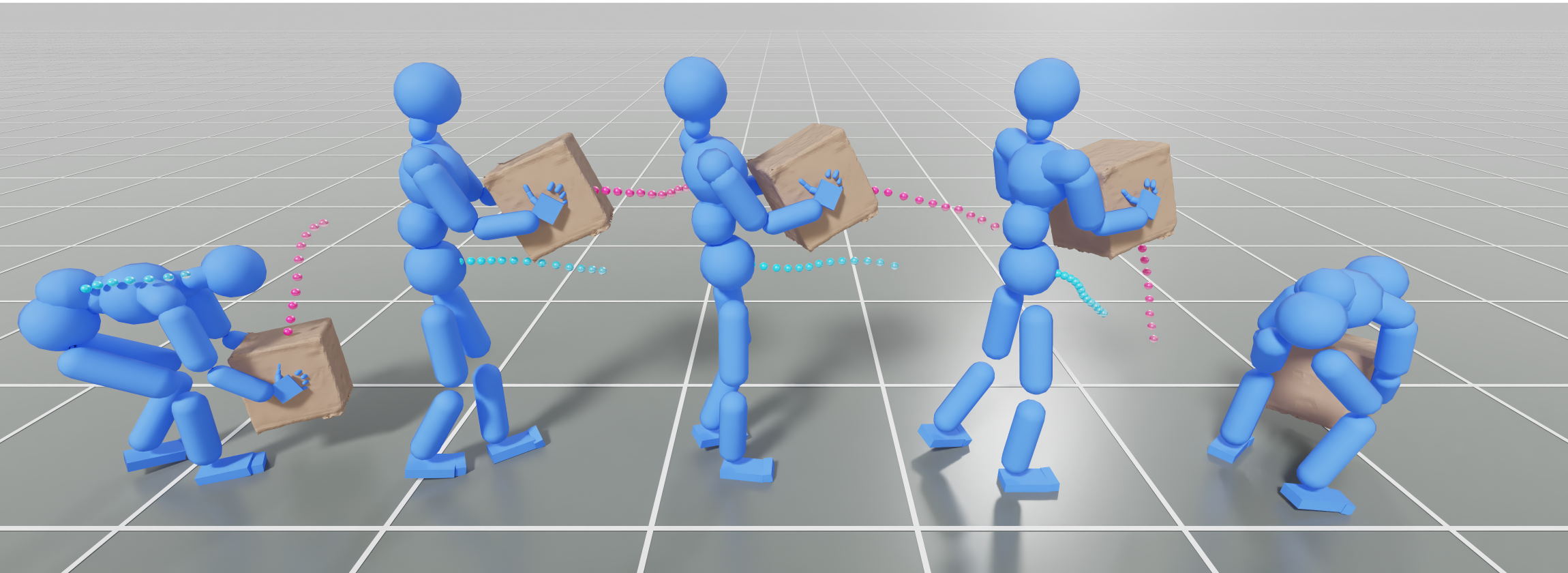}
    \caption{\textbf{Receding-horizon trajectory planning.}
    The planner predicts a 30-step {\color{purple}{object}} and {\color{cornflowerblue}{humanoid}} root trajectory (dotted curves), which guides the action generator and is refreshed every 8 control steps.
    }
    \label{fig:planner}
\end{figure}

\paragraph{Trajectory planner.}
The planner is defined as
\begin{equation}
\begin{aligned}
    \mathbf{P}_t
    &\sim
    \pi_{\mathrm P}
    \left(\,\cdot\mid\mathbf{q}_{\leq t},y,g\right),
\end{aligned}
\label{eq:planner}
\end{equation}
where $\mathbf{q}_{\leq t}=(\mathbf{q}_0,\ldots,\mathbf{q}_t)$ is the history of the object and humanoid-root poses $\mathbf{q}_i$, taken from $\mathbf{s}_i$.
Restricting the planner’s state input to the compact representation of object and humanoid-root poses allows us to incorporate motion history from the full episode while keeping computation lightweight.
$\mathbf{P}_t$ is a sequence of future object and humanoid-root poses over a horizon of $T_{\mathrm P}$ steps. %
The object trajectory specifies the intended object motion, and the humanoid-root trajectory specifies the coarse locomotion that this motion requires.
An illustration is shown in Fig.~\ref{fig:planner}.

\begin{figure}[!t]
    \centering
    \includegraphics[width=0.5\textwidth]{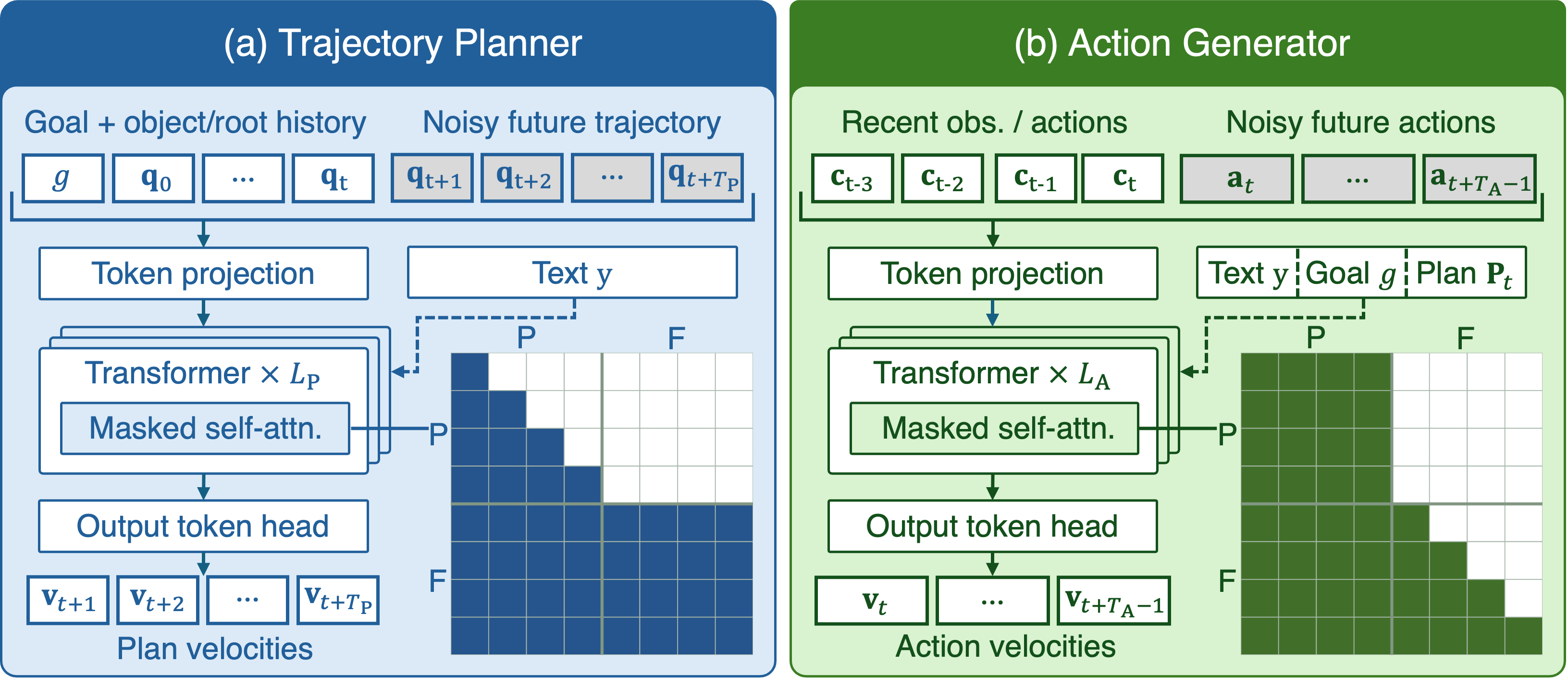}
    \caption{
    \textbf{Architectures of the trajectory planner and action generator.}
    Gray tokens denote noisy inputs.
    In the attention masks, P/F denote prefix/future tokens, and colored/white cells indicate allowed/blocked attention; rows are queries and columns are keys.
    }
    \label{fig:model_arch}
\end{figure}

As shown in Fig.~\ref{fig:model_arch} (a), the input to the planner transformer consists of the token of terminal object pose $g$, the tokens of object--root history $\mathbf{q}_{\leq t}$, and the noised trajectory tokens to be denoised.
The goal and history tokens form a causal prefix that does not attend to future trajectory tokens, enabling KV caching for efficient planner inference.
Each noised trajectory token attends to the prefix and other trajectory tokens bidirectionally, 
so that later poses inform earlier ones and the plan stays consistent along the horizon.
The instruction $y$ is encoded by a frozen CLIP text encoder and conditions every transformer block through adaptive normalization.

\paragraph{Whole-body action generator.}
The action generator is defined as
\begin{equation}
\begin{aligned}
    \mathbf{A}_t
    &\sim
    \pi_{\mathrm A}
    \left(\,\cdot\mid\mathbf{C}_t,\mathbf{P}_t,y,g\right),
\end{aligned}
\label{eq:action_generator}
\end{equation}
where $\mathbf{C}_t$ is the control context of the $K$ latest observations and executed actions,
\begin{equation}
    \mathbf{C}_t
    =
    \left[
        \mathbf{c}_i
    \right]_{i=t-K+1}^{t}
    =
    \left[
        \left(
            \mathbf{o}_i,
            \mathbf{a}_{i-1}
        \right)
    \right]_{i=t-K+1}^{t}.
    \label{eq:action_context}
\end{equation}
Pairing each executed action with the observation that follows it shows the model the effect of its own actions.
The shared heading frame for the pose history, terminal object pose, and predicted trajectories is anchored at the humanoid root at step $t-K+1$, the first step of the control context $\mathbf{C}_t$.
During training, the action generator is supervised with sequences $\mathbf{A}_t$ of $T_{\mathrm A}$ future 153-dimensional joint-target actions.

As illustrated by Fig.~\ref{fig:model_arch} (b), 
we implement the action generator as a transformer over context and future-action tokens.
Each context pair $\mathbf{c}_i$ is embedded as one context token, and each noisy future action is embedded as one future-action token.
The interaction features $\boldsymbol{\phi}$ are encoded by a shared MLP and concatenated with the other inputs before token projection.
The context tokens attend to one another in both directions and never to the future-action tokens.
Each future-action token attends to all context tokens and to the earlier future-action tokens only.
This causal structure allows us to generate only the action tokens needed for execution, avoiding full-sequence generation and accelerating inference.
The instruction $y$, the terminal object pose $g$, and the plan $\mathbf{P}_t$ condition every transformer block through adaptive normalization.

\begin{figure*}[!t]
    \centering
    \includegraphics[width=0.95\textwidth]{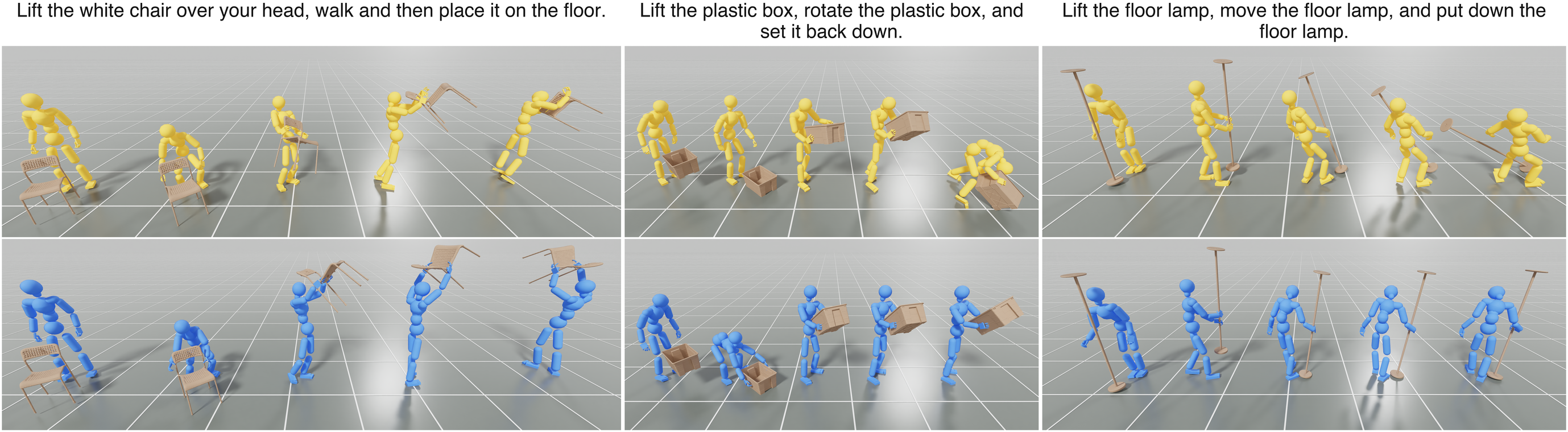}
    \caption{
    \textbf{Qualitative effect of on-policy post-tuning.}
    Frames are ordered left to right; the top row shows \textcolor{intermimicyellow}{Flow-BC} and the bottom row shows \textcolor{blue!70}{Flow-BC + FT}.
    While Flow-BC captures the intended behavior, small execution errors accumulate into unstable whole-body compensation or loss of object control, evident during overhead chair transport, box rotation, and floor-lamp relocation.
    On-policy post-tuning preserves balance and maintains stable object contact through these challenging phases, enabling execution to proceed toward the target placement.
    }
    \label{fig:bcvsrl}
\end{figure*}

\paragraph{Model training.}
The successful rollouts collected in Sec.~\ref{sec:tracking} supervise both the planner and the action generator via behavior cloning.
We train both models with the conditional flow-matching loss~\citep{lipman2022flow}.
For a target sequence $\mathbf{x}$ and model-specific conditioning $\mathbf{c}$, we sample
$\mathbf{z}\sim\mathcal{N}(\mathbf{0},\mathbf{I})$ and
$\tau\sim\mathcal{U}[0,1]$, set
$\mathbf{x}_{\tau}=(1-\tau)\mathbf{z}+\tau\mathbf{x}$, and minimize
\begin{equation}
    \mathcal{L}_{\mathrm{FM}}
    =
    \mathbb{E}
    \left[
        \left\|
            \mathbf{v}_{\theta}
            \!\left(
                \mathbf{x}_{\tau},
                \tau;
                \mathbf{c},y,g
            \right)
            -
            \left(
                \mathbf{x}-\mathbf{z}
            \right)
        \right\|_2^2
    \right],
    \label{eq:flow_matching_loss}
\end{equation}
where $\mathbf{v}_{\theta}$ is the velocity field predicted by the transformer, and $y$ and $g$ condition both models.
For a training example cut from a rollout at a step $t$, the planner has target $\mathbf{x}=(\mathbf{q}_{t+1},\ldots,\mathbf{q}_{t+T_{\mathrm P}})$ and conditioning $\mathbf{c}=\mathbf{q}_{\leq t}$, and the action generator has target $\mathbf{x}=(\mathbf{a}_t,\ldots,\mathbf{a}_{t+T_{\mathrm A}-1})$ and conditioning $\mathbf{c}=(\mathbf{C}_t,\mathbf{P}_t)$.

The two models are trained jointly, but we apply stop-gradient to the trajectory \(\mathbf{P}_t\) when conditioning the action generator.
Early in training this trajectory \(\mathbf{P}_t\) is mostly the one from the expert rollout, and it is progressively replaced by trajectories sampled from the planner to reduce the distribution discrepancy between training and inference.

\paragraph{Model inference.}

At inference, each model draws $\mathbf{z}\sim\mathcal{N}(\mathbf{0},\mathbf{I})$ and samples its output by integrating $\mathrm{d}\mathbf{x}_{\tau}/\mathrm{d}\tau=\mathbf{v}_{\theta}(\mathbf{x}_{\tau},\tau;\mathbf{c},y,g)$ from $\mathbf{x}_0=\mathbf{z}$ to $\tau=1$.
The planner is re-run at fixed intervals, and its predicted trajectory is held between updates to give the action generator a consistent trajectory over each interval.
As rollout proceeds, we remove the elapsed poses from the trajectory and re-express the remainder in the shared heading frame.
We continue to denote this remaining trajectory by $\mathbf{P}_t$.
At every control step, the action generator samples a single action token conditioned on the latest context $\mathbf{C}_t$, the remaining trajectory $\mathbf{P}_t$, the instruction $y$, and the terminal object pose $g$.
The resulting action is executed as $\mathbf{a}_t$.

\subsection{On-Policy Post-Tuning}
\label{sec:flow_rl}

During closed-loop execution, contact errors can drive the system beyond the states covered by expert rollouts, where the behavior-cloned action generator may fail to recover (Fig.~\ref{fig:bcvsrl}).
We therefore post-tune the action generator as shown in Stage III of Fig.~\ref{fig:method}.
Since the language instruction and terminal object goal do not provide per-step spatial targets for post-tuning, we reuse the planner's predicted object and humanoid-root trajectories as task supervision.
The planner is kept frozen during post-tuning, so that these predictions provide a consistent training signal while the action generator adapts to its on-policy rollouts.

\paragraph{Post-tuning reward.}
During post-tuning, each episode is initialized from a state sampled from a retained Stage~I rollout (Sec.~\ref{sec:tracking}).
To retain the interaction behavior learned offline while improving physical execution, we combine low-rank adaptation (LoRA)~\citep{hu2021lora} with velocity-field regularization.
LoRA restricts the trainable weight updates to a low-rank form while keeping the pretrained generator weights frozen, and the regularizer penalizes deviations from the frozen behavior-cloned velocity field to discourage drift.

We adopt the Flow-SDE formulation~\citep{chen2025pirl}, which converts deterministic flow sampling into a stochastic process with tractable transition likelihoods, enabling PPO-based policy optimization.
The LoRA-adapted action generator serves as the actor, while a separate MLP critic estimates the value function for PPO updates.
The per-step reward $r_t$ uses no per-frame kinematic reference and combines motion naturalness, task completion, and task quality,
\begin{equation}
\begin{aligned}
    r_t
    &=
    w_{\mathrm{motion}}r^{\mathrm{motion}}_t
    +b_{\mathrm{goal}}\mathbf{1}_{\mathrm{goal}}(t)
    +w_{\mathrm{task}}r^{\mathrm{task}}_t,
\end{aligned}
\label{eq:rl_reward}
\end{equation}
where $r^{\mathrm{motion}}_t$ measures the quality using a frozen text-conditioned score-matching motion prior~\citep{mu2025smp}, which discourages mechanically effective but visually unnatural motion.
The prior evaluates a short window of humanoid kinematics and does not observe the object, leaving manipulation progress to the task reward.
The indicator $\mathbf{1}_{\mathrm{goal}}(t)$ is one at the first step satisfying the terminal-pose dwell criterion and zero otherwise, giving the character a one-time task finishing reward.

The task quality is defined as 
\begin{align}
        r^{\mathrm{task}}_t
    &=
    \exp\!\left[
        -\left(
            \lambda_{\mathrm{obj}}
             m^{\mathrm{obj}}(t)^2
            +\lambda_{\mathrm{root}}
             m^{\mathrm{root}}(t)^2
            +E^{\mathrm{int}}_t
        \right)
    \right],
\end{align}
where $m^{\mathrm{obj}}(t)$ and $m^{\mathrm{root}}(t)$ are the trajectory-following errors for the object and humanoid-root, respectively. 
$E^{\mathrm{int}}_t$ is the interaction error. 
A large error in either trajectory following or the interaction quality strongly suppresses the task reward.
We introduce them in detail below.

\textbf{Trajectory-following errors.}
To guide the action generator along the frozen planner’s trajectories, we measure discrepancies between simulated and predicted object and humanoid-root poses.
\begin{align}
m^{\ell}(t)
= \left\{\begin{array}{ll}
        \frac{1}{|\mathcal{V}_t|}
        \sum_{j\in\mathcal{V}_t}
        d^{\ell}\!\left(
        \mathbf{q}^{\ell}_t,
        \tilde{\mathbf{q}}^{\ell}_{t\mid j}
        \right),
         & \mathcal{V}_t\neq\varnothing, \\
        0 & \mathcal{V}_t=\varnothing.
    \end{array}\right.
\end{align}
Here, \(\tilde{\mathbf{q}}^{\ell}_{t\mid j}\) denotes the pose predicted for step \(t\) by the planner produced at step \(j\), and \(d^{\ell}\) measures its discrepancy from the corresponding simulated pose $\mathbf{q}^{\ell}_t$.
We store each plan’s predicted trajectories. 
\(\mathcal{V}_t\) is the set of planning steps \(j\) whose predicted horizons include the current step \(t\). 
Several earlier plans can predict poses for the same control step. Averaging their discrepancies reduces sensitivity to an individual plan sample and measures consistency with the planner’s recent predictions.

\textbf{Simulation-based interaction quality.}
Trajectory following constrains the object and humanoid-root motion but does not determine how the hands establish and maintain contact.
We therefore use the interaction cost $E^{\mathrm{int}}_t$ to evaluate the interaction realized in simulation.
It combines physical hand--object contact, the geometry-conditioned grasp cost, palm alignment, hold stability, and the energy costs on joint and object accelerations and contact forces introduced in Sec.~\ref{sec:tracking}.
Because Stage~III has no kinematic reference, we omit the reference contact labels and the reference contact-anchor term.
For the geometry-conditioned grasp cost, the interaction anchor $\mathbf{r}$ is instead chosen as the object-surface point nearest to the current simulated hand.
The reference-derived gate $\eta$ is likewise replaced by activation based on simulated hand--object proximity.

\begin{table*}[!th]
\centering
\caption{\textbf{Contact-rich interaction tracking on OMOMO~\cite{li2023object}.}
Top: controlled ablation of the proposed interaction objectives under the InterMimic~\cite{xu2025intermimic} training protocol.
Bottom: performance of the unified tracking policy trained on the full retargeted dataset. $^\dagger$Results from our Isaac Lab reimplementation based on the official InterMimic code, retrained under the same subject-specific protocol.}
\label{tab:reward_ablation}
\begin{tabular}{lccccc}
\toprule
Method & SR (\%)\,$\uparrow$ & Time (fr)\,$\uparrow$ & Human (cm)\,$\downarrow$ & Object (cm)\,$\downarrow$ & Jitter (m/s$^2$)\,$\downarrow$ \\
\midrule
\rowcolor{gray!20} \multicolumn{6}{c}{\bf Controlled comparison (3 subject-specific policies per method)} \\
InterMimic~\cite{xu2025intermimic}$^\dagger$ & 53.23 & 89.5 & 7.84 & 12.89 & 18.77 \\
Ours w/o grasp cost     & 59.20 & 95.4 & 8.36 & 13.83 & 18.43 \\
Ours w/o palm alignment      & 61.25 & 96.5 & 8.13 & 13.62 & 17.39 \\
Ours w/o contact anchor    & 60.18 & 95.5 & 8.16 & 14.06 & 18.37 \\
Ours w/o hold stability      & 59.30 & 96.7 & 8.51 & 14.16 & 18.39 \\
Ours w/o gated reference tracking   & 59.88 & 93.4 & \textbf{7.49} & \textbf{12.32} & 17.64 \\
\textbf{Ours} & \textbf{64.29} & \textbf{97.0} & 8.17 & 13.55 & \textbf{16.90} \\
\midrule
\rowcolor{gray!20} \multicolumn{6}{c}{\bf Unified tracking policy} \\
Ours (unified) & 76.50 & 148.6 & 7.50 & 15.13 & 12.41 \\
\bottomrule
\end{tabular}
\end{table*}

\paragraph{Reference-free terminations.}
Unlike the tracking stage, Stage III has no full-body kinematic reference, so failures must be detected from the planner and simulated states themselves.
We use \emph{plan divergence}, measured by $m_{\mathrm{obj}}(t)$ and $m_{\mathrm{root}}(t)$, to detect large rollout failures such as dropped objects or motion that departs from the planned trajectory.

However, a small plan error does not necessarily indicate successful execution: when the planner requests only a small object movement, even a stationary object can remain within the error tolerance.
This limitation arises because the plan divergence measures error on an \emph{absolute} scale, without accounting for the amount of motion requested.
We therefore introduce a complementary \emph{object-progress criterion}, which measures the \emph{relative} fraction of the planned object displacement that has actually been achieved.
Together, the two criteria detect both large deviations from the plan and insufficient progress along it.
A humanoid fall also terminates the episode, whereas successful completion requires the object to remain within the terminal-pose tolerance for the prescribed dwell period.
Detailed definitions are provided in the appendix.

\section{Experiments}
\label{sec:experiments}
We evaluate LYRIC on OMOMO~\citep{li2023object}.
Our evaluations assess task success, instruction adherence, and the quality of the simulated character motion.
\subsection{Implementation Details}
\label{sec:exp_detail}
We implement all models in PyTorch and use Isaac Lab for simulation.
The control loop runs at $30$\,Hz, while the simulator steps at $120$\,Hz.
The actor and critic of the unified tracking policy are MLPs with hidden widths $(4096,4096,2048)$ and ReLU activations.
The subject-specific tracking policies evaluated in Sec.~\ref{sec:exp_tracking} use smaller actor and critic MLPs with hidden widths $(1024,1024,512)$.
For Stage~I rollout collection, we generate 100 rollouts per demonstration.

In Stage~II, the planner and action generator have $L_{\mathrm P}=6$ and $L_{\mathrm A}=12$ transformer blocks, with model widths $192$ and $768$, respectively.
The planner predicts object and humanoid-root trajectories over $T_{\mathrm P}=30$ control steps and refreshes them every eight control steps.
The action generator conditions on $K=4$ observation--action pairs and is trained with sequences of $T_{\mathrm A}=8$ future actions.
We jointly train the two models with AdamW for one million updates on eight NVIDIA H100 GPUs, using a global batch size of $2048$ and a peak learning rate of $2\times10^{-4}$.
Unless otherwise stated, both models sample their outputs using five Euler integration steps with deterministic flow ODE integration at inference.

Post-tuning uses LoRA with rank $16$ and scaling parameter $\alpha=32$.
The actor learning rate is $10^{-4}$, and the velocity-field regularizer has weight $10^{-3}$.
We set a fixed budget of $2000$ PPO iterations on eight NVIDIA H100 GPUs.
Complete simulation settings, interaction features, reward definitions, and training hyperparameters are provided in the appendix.

\subsection{Dataset and Evaluation Metrics}
\label{sec:exp_setup}

\paragraph{Dataset.}
OMOMO~\citep{li2023object} contains 4,890 motion-capture sequences of 17 subjects interacting with 13 objects.
All simulated rollout evaluations initialize each episode from frame 0 of the corresponding sequence.

\paragraph{Evaluation metrics.}
We evaluate reference-tracking performance and language-driven interaction control, together with physical artifacts and temporal motion quality.

{\textbf{Tracking policy metrics.}}
Following InterMimic~\cite{xu2025intermimic}, we report tracking success rate, average rollout duration, character tracking error, and object tracking error.
\emph{Tracking success rate} is the percentage of sequences completed without triggering any of the failure conditions defined in Sec.~\ref{sec:tracking}.
\emph{Time} is the average number of executed control frames.
\emph{Human} and \emph{Object} measure character and object reference-tracking errors in centimeters.
Following MotionBricks~\cite{wang2026motionbricks}, we additionally report \emph{Jitter} to measure the motion smoothness.

{\textbf{Text-conditioned controller metrics.}}
We report task success, rollout duration, humanoid-motion alignment with the instruction, and motion diversity.
\emph{Success rate} is computed under the repeated-attempt protocol described in Sec.~\ref{sec:exp_generation}.
A controller rollout is successful if it satisfies the task-completion condition before triggering an applicable failure termination defined in Sec.~\ref{sec:flow_rl}.
\emph{Time} is the average number of control frames executed per rollout.

\begin{figure}[t]
    \centering
    \setlength{\tabcolsep}{0.8pt}
    \begin{tabular}{c}
        \includegraphics[width=0.99\linewidth]{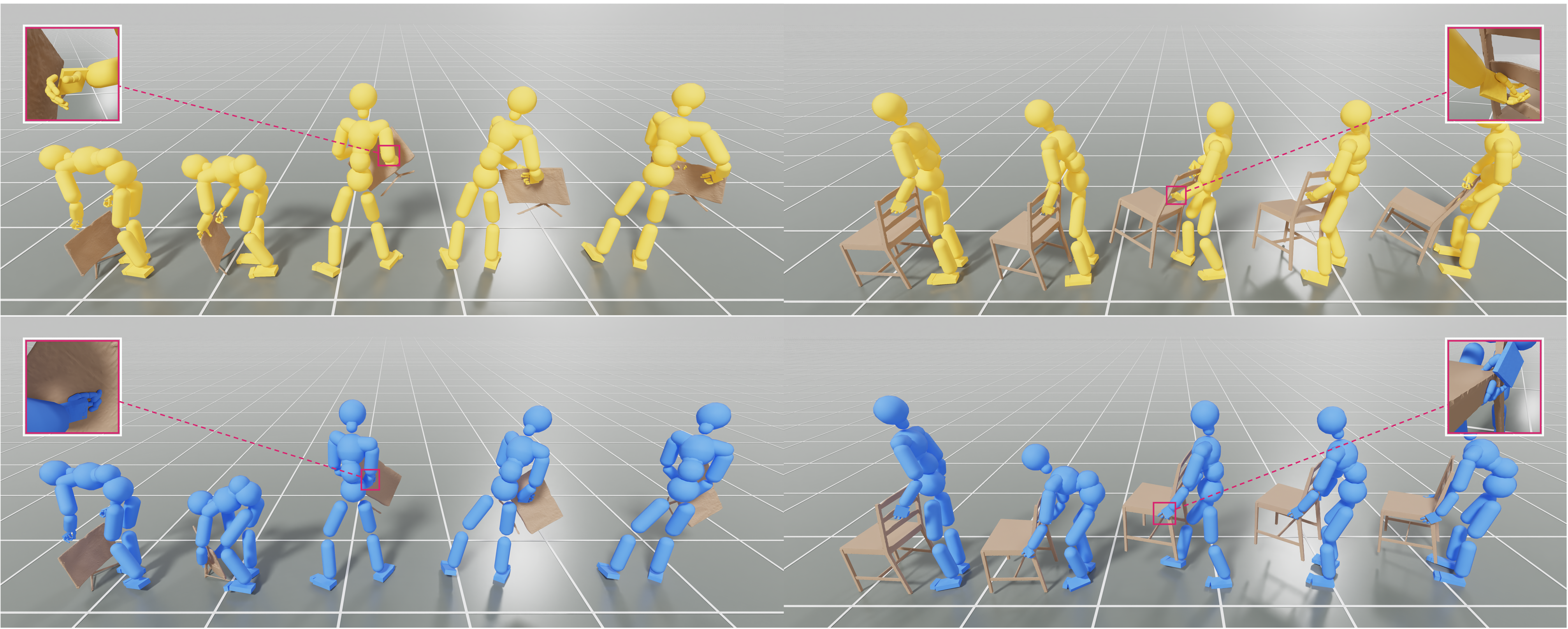} \\
        (a) More natural hand--object interaction. \\
        \includegraphics[width=0.99\linewidth]{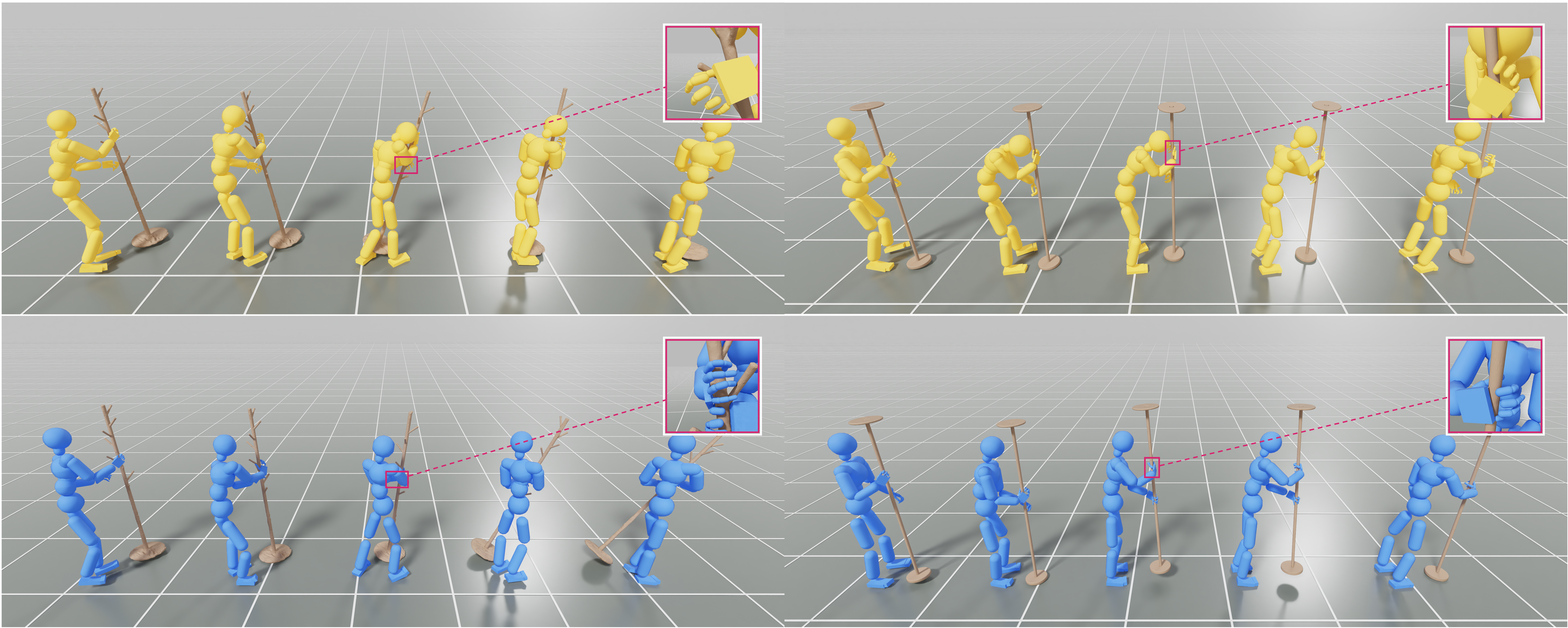} \\
            (b) Avoiding compensatory motion caused by poor grasp quality. \\
        \includegraphics[width=0.99\linewidth]{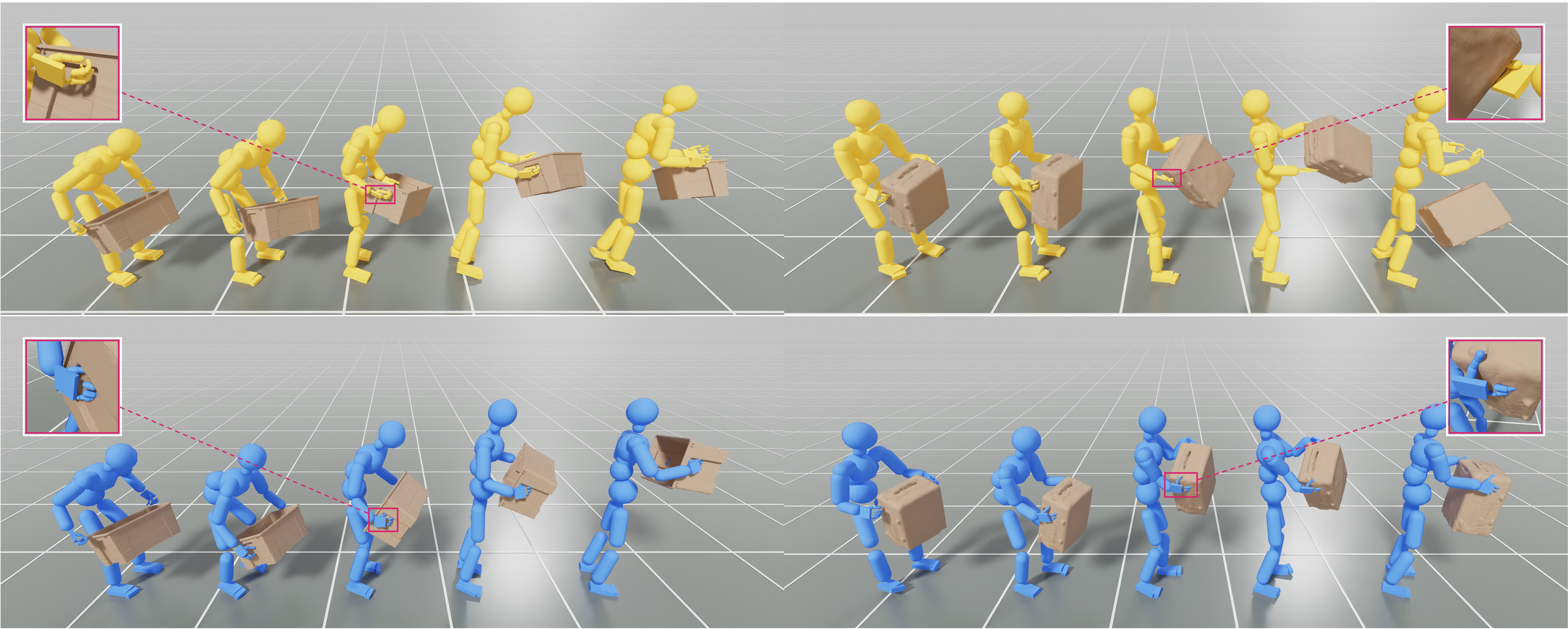} \\
        (c) Stable execution during challenging manipulation. \\
        
    \end{tabular}
    \caption{\textbf{Qualitative comparison between our tracking policy and InterMimic.}
    Top: \textcolor{intermimicyellow}{InterMimic}; bottom: \textcolor{blue!70}{our stage~I tracking policy}.
    Our geometry-conditioned tracking policy produces more natural and stable interactions, while InterMimic often exhibits less suitable grasping, compensatory body motion, or failure to maintain the interaction.
    }
    \label{fig:tracking_compare}
\end{figure}

We assess the alignment between the instruction and the realized humanoid motion using a learned text--motion evaluator.
We train the evaluator on retargeted OMOMO using the model and training protocol in prior work~\citep{guo2022generating}.
Following MARDM~\citep{meng2025rethinking}, the evaluator uses an essential-motion representation comprising root motion and the positions of 21 body joints.
It does not observe object or finger motion, so its scores characterize only the humanoid-motion component of instruction alignment.
Complete evaluator training and scoring details are provided in Appendix~\ref{app:evaluator}.
The evaluator provides three metrics.
\emph{R@1} and \emph{R@3} measure the fractions of motions for which the paired instruction is retrieved among the top one and top three text candidates, respectively.
\emph{MM-Dist} measures the mean Euclidean distance between paired text and motion embeddings.
\emph{Diversity} measures the mean Euclidean distance between sampled pairs of motion embeddings and is interpreted relative to the reference-motion value rather than maximized independently.

{\textbf{Physical and temporal motion-quality metrics.}}
Task success, duration, and instruction alignment do not directly measure physical artifacts or temporal irregularities. We therefore report separate metrics for these two aspects of motion quality.
We compute Penetrate, Float, and Skate using the public HumoS~\citep{tripathi2024humos} implementation.
\emph{Penetrate} measures the depth of the lowest body vertex below the ground tolerance, while \emph{Float} measures the clearance of the lowest body vertex above it.
\emph{Skate} measures the horizontal displacement of ground-contacting feet across adjacent frames.
Following PhysDiff~\citep{yuan2023physdiff}, \emph{Phys-Err} aggregates these three physical artifacts as the sum of Penetrate, Float, and Skate.
All four quantities are reported in millimeters.
\emph{Jerk} measures third-order temporal variation in $\mathrm{m}/\mathrm{s}^{3}$.
We use the public implementation released with prior work~\citep{tanke2021intention} to compute \emph{NDMS}, which measures the similarity of local joint-velocity windows to real motion.

\subsection{Contact-Rich Interaction Tracking}
\label{sec:exp_tracking}
\paragraph{Comparison setup and evaluation protocol.}
InterMimic~\citep{xu2025intermimic} is the closest baseline in terms of training physics-based policies to track paired humanoid and object trajectories.
To ensure fair comparison, following its subject-specific protocol, we train one tracking policy per subject on the original OMOMO data without retargeting.
We reimplement its tracking pipeline in Isaac Lab, with all ablation variants matching the policy capacity, base tracking formulation, and training setup, and differing only in the hand-interaction components introduced in Sec.~\ref{sec:tracking}.
Because evaluating every variant over all 17 subjects would require training 17 policies for each ablation variant, we conduct the ablation on a three-subject subset of 1,022 sequences that preserves the object coverage of the full dataset.

We additionally evaluate five variants of our complete tracking objective, each removing one component introduced in Sec.~\ref{sec:tracking} while retaining the others.
\emph{Ours w/o grasp cost} removes the geometry-conditioned grasp cost $E_t^{\mathrm{grasp}}$, including both its enclosure and support modes.
\emph{Ours w/o palm alignment} removes the regularizer that encourages the palm to face the local object surface.
\emph{Ours w/o contact anchor} removes the penalty on deviation from the reference hand-contact region, while retaining the interaction anchor used to compute local grasp geometry.
\emph{Ours w/o hold stability} removes the penalties on sliding finger contacts and the loss of established contacts.
\emph{Ours w/o gated reference tracking} disables the contact-dependent relaxation of reference tracking for the hand and supporting arm chain, while retaining the grasp cost and its reference-contact gate.

Separately, for the unified tracking policy used to collect Stage~II training data, we retarget the complete dataset to a canonical SMPL-X embodiment using our interaction-preserving adaptation of OmniRetarget~\citep{yang2025omniretarget} and train one unified policy jointly over all subjects and objects with object-identity one-hot conditioning.
Because this setting changes both the training data and the policy configuration, we report it separately from the controlled comparison.

\begin{table*}[t]
\centering
\caption{\textbf{Text-conditioned character control on OMOMO.}
Flow-BC denotes our controller with a trajectory planner and an action generator, trained by behavior cloning as described in Sec.~\ref{sec:flow}; Flow-BC (w/o planner) removes the intermediate trajectory planner and conditions the action generator directly on the 9-D terminal object goal.
Flow-BC + FT denotes LYRIC, our final controller after the proposed RL post-tuning in Sec.~\ref{sec:flow_rl}.
The \textit{Mocap} row reports the corresponding metrics computed on the ground-truth motion-capture sequences.
}
\label{tab:flow_policy}
\begin{adjustbox}{max width=\textwidth}
\begin{tabular}{lcc cccc cccccc}
\toprule
& \multicolumn{2}{c}{\bf Task}
& \multicolumn{4}{c}{\bf Semantic}
& \multicolumn{6}{c}{\bf Motion quality} \\
\cmidrule(lr){2-3}\cmidrule(lr){4-7}\cmidrule(lr){8-13}
Method
& SR (\%)\,$\uparrow$
& Time (fr)
& R@1\,$\uparrow$
& R@3\,$\uparrow$
& MM-Dist\,$\downarrow$
& Div.\,$\rightarrow$
& Penet.\,$\downarrow$
& Float\,$\downarrow$
& Skate\,$\downarrow$
& Phys-Err\,$\downarrow$
& Jerk\,$\downarrow$
& NDMS\,$\uparrow$ \\
\midrule

\rowcolor{gray!20}
\multicolumn{13}{c}{\bf Train split (3485 sequences)} \\

\textit{Mocap (reference)}
& -- & --
& \textit{0.669}
& \textit{0.963}
& \textit{1.60}
& \textit{15.76}
& \textit{15.62}
& \textit{6.56}
& \textit{7.45}
& \textit{29.27}
& \textit{38.9}
& \textit{0.579} \\

Flow-BC (w/o planner)
& 83.01
& 187.8
& 0.354
& 0.639
& 5.57
& 13.93
& 0.07
& 13.04
& \textbf{6.93}
& \textbf{20.05}
& \textbf{269.9}
& 0.427 \\

Flow-BC
& 80.32
& 128.3
& \textbf{0.390}
& \textbf{0.709}
& 4.62
& 14.69
& 0.03
& \textbf{12.98}
& 7.27
& 20.27
& 290.2
& 0.439 \\

\textbf{Flow-BC + FT (LYRIC)}
& \textbf{91.33}
& 135.2
& 0.374
& 0.698
& \textbf{4.61}
& \textbf{15.08}
& \textbf{0.02}
& 13.25
& 7.19
& 20.46
& 281.7
& \textbf{0.452} \\

\midrule

\rowcolor{gray!20}
\multicolumn{13}{c}{\bf Held-out split (267 sequences)} \\

\textit{Mocap (reference)}
& -- & --
& \textit{0.488}
& \textit{0.853}
& \textit{2.91}
& --
& \textit{12.76}
& \textit{8.95}
& \textit{8.06}
& \textit{29.77}
& \textit{41.0}
& \textit{0.570} \\

Flow-BC (w/o planner)
& 79.03
& 156.7
& 0.323
& 0.604
& 5.84
& --
& 0.03
& 7.35
& 7.00
& 14.38
& 269.6
& 0.439 \\

Flow-BC
& 74.16
& 119.8
& 0.347
& 0.626
& 5.56
& --
& \textbf{0.00}
& 8.12
& 7.00
& 15.12
& 281.2
& 0.440 \\

\textbf{Flow-BC + FT (LYRIC)}
& \textbf{90.26}
& 148.8
& \textbf{0.361}
& \textbf{0.666}
& \textbf{4.94}
& --
& 0.02
& \textbf{7.08}
& \textbf{6.75}
& \textbf{13.85}
& \textbf{256.7}
& \textbf{0.451} \\

\bottomrule
\end{tabular}
\end{adjustbox}
\end{table*}

\paragraph{Quantitative evaluation.}
Table~\ref{tab:reward_ablation} evaluates the proposed hand-interaction components under the controlled subject-specific training protocol.
Under matched policy capacity and training settings, our complete objective achieves a success rate of $64.29\%$, compared with $53.23\%$ for InterMimic, an absolute improvement of 11.06 percentage points.
Removing any proposed component reduces success to between $59.20\%$ and $61.25\%$.
The largest reductions occur without the geometry-conditioned grasp cost and hold stability, which lower success to $59.20\%$ and $59.30\%$, respectively.
The complete objective also produces the longest average rollout and the lowest jitter among the controlled variants.
Together, these results indicate that the proposed terms provide complementary guidance for contact acquisition, geometry-adaptive grasp formation, and stable contact maintenance.

Importantly, success and tracking error capture different objectives: successful physical interaction can require deliberate deviation from the kinematic reference.
This is evident when gated reference tracking is removed, which achieves the lowest human and object errors ($7.49$ cm and $12.32$ cm) but reduces success from $64.29\%$ to $59.88\%$.
Strictly tracking the captured arm and hand motion limits the local adjustments needed to accommodate the simulated object geometry and establish viable contacts.
Our complete objective therefore deliberately relaxes reference fidelity where necessary, yielding substantially higher interaction success with only modest increases in tracking error.

The bottom block reports the unified tracking policy used to collect Stage~II training trajectories. 
This single policy achieves a $76.50\%$ success rate and an average rollout duration of $148.6$ frames on the complete retargeted OMOMO dataset. 
Its successful rollouts provide abundant and high-quality trajectories across the dataset for training the generative controller, without requiring separate subject-specific policies or a subsequent distillation stage.

\paragraph{Qualitative comparison.}
Figure~\ref{fig:tracking_compare} further illustrates the effect of the proposed hand-interaction costs.
In Fig.~\ref{fig:tracking_compare} (a), both methods complete the interaction, but our policy produces cleaner and more animation-friendly hand--object motion, with contact configurations that better match the local object geometry.
In Fig.~\ref{fig:tracking_compare} (b), the weaker grasp quality of InterMimic leads to unnatural whole-body compensation to keep the object in place, while our method maintains stable contact and more natural character motion throughout the sequence.
In Fig.~\ref{fig:tracking_compare} (c), the contact failure becomes more severe: InterMimic loses the interaction entirely, whereas our tracker remains stable and completes the task successfully.
These examples show that the benefit of our geometry-conditioned interaction rewards is not only higher task success, but also improved motion quality and more plausible hand--object behavior.
We have more video results for our unified tracking policy on the supplementary webpage.

\begin{figure*}[!t]
    \centering
    \includegraphics[width=0.95\textwidth]{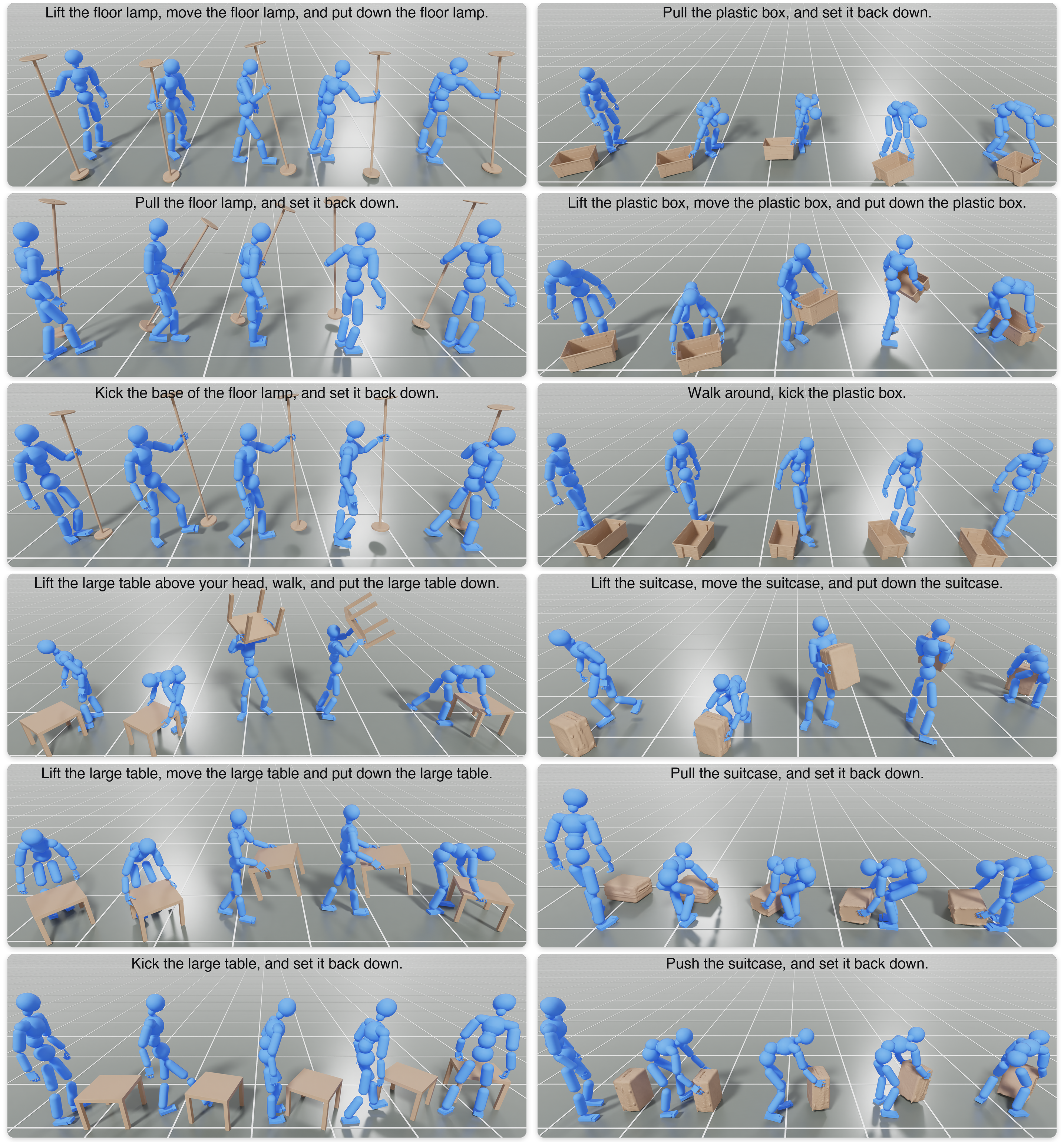}
    \caption{\textbf{Gallery of contact-rich whole-body object interactions driven by language instructions.}
    Representative rollouts of LYRIC, with frames in each sequence ordered from left to right. See the supplementary webpage for additional video results.
    }
    \label{fig:gallery}
\end{figure*}

\subsection{Text-Conditioned Character Control}
\label{sec:exp_generation}
\paragraph{Data and evaluation protocol.}
We collect a clean dataset of 3,752 successful trajectories using the unified tracking policy with observation noise disabled.
Following InterMimic~\citep{xu2025intermimic}, we use this dataset for Stage~III episode initialization and evaluate the generative controller on these retained sequences rather than the full OMOMO dataset.
Each evaluation episode of the text-conditioned controller runs for at most 1,000 control steps.
Following existing work~\citep{xu2025intermimic,xu2026interprior}, \emph{generative controller success rate} is reported per sequence over repeated attempts under a fixed global episode budget and a sequence is counted as solved if at least one attempt succeeds.
Of the 3,752 retained trajectories, 267 are from subject~14 and are reserved for held-out evaluation, following InterMimic~\citep{xu2025intermimic}.
Our evaluation protocol requires more than 300 valid motion embeddings to compute Diversity.
Since the held-out split contains only 267 trajectories, we omit Diversity on this split.

Table~\ref{tab:flow_policy} summarizes the quantitative results for text-conditioned character control. We discuss the effectiveness of the trajectory planner and planner-guided on-policy post-tuning below.

\paragraph{Effectiveness of the trajectory planner.}
To evaluate the contribution of the planner $\pi_{\mathrm P}$, we train \emph{Flow-BC (w/o planner)}, which removes the trajectory planner and conditions the action generator directly on the 9-D terminal object goal instead of the object and humanoid-root trajectories predicted over the next 30 steps.
All other Stage-II training settings are kept unchanged.
Evaluation uses the same protocol except that the \emph{Flow-BC (w/o planner)} variant omits the plan-divergence and object-progress terminations.

Since removing the planner disables the plan-divergence and object-progress termination terms, the controller has more opportunity to complete placement before an episode terminates.
Under these more permissive termination conditions, we observe a modest increase in placement success.
The longer average rollout durations on both the training and held-out splits also reflect the additional execution time afforded by removing these termination terms.
Consequently, the planner-free variant's success rates are not directly comparable with those of variants that retain these termination terms.
Meanwhile, removing the planner substantially weakens semantic alignment.
On the training split, R@3 decreases from $0.709$ to $0.639$ and MM-Dist increases from $4.62$ to $5.57$.
The same trends hold on the held-out split.
These results support the planner's role in preserving text-specified interaction behavior: its predicted object and humanoid-root trajectories guide how the character manipulates the object, rather than only specifying where the object should be placed.

\begin{table*}[t]
\centering
\caption{\textbf{Comparison with kinematic-planner baselines on OMOMO.}
1-shot kin. planner generates a complete text-conditioned kinematic character--object motion sequence once, which a closed-loop tracking policy then executes.
Replan. kin. planner uses CLoSD-style~\citep{tevet2024closd} closed-loop replanning, repeatedly generating a short
kinematic horizon from recent simulated motion and executing a prefix with
the tracker.
1-shot kin. planner + FT and Replan. kin. planner + FT keep the corresponding generator frozen and further
fine-tune the tracker against its generated references using PPO.
Time denotes average rollout duration in control frames.
}
\label{tab:flow_baseline_comparison}
{\setlength{\tabcolsep}{3pt}
\begin{tabular}{l@{\hspace{2pt}}cc cccc cccccc}
\toprule
& \multicolumn{2}{c}{\bf Task}
& \multicolumn{4}{c}{\bf Semantic}
& \multicolumn{6}{c}{\bf Motion quality} \\
\cmidrule(lr){2-3}
\cmidrule(lr){4-7}
\cmidrule(lr){8-13}
Method
& SR (\%)\,$\uparrow$
& Time (fr)
& R@1\,$\uparrow$
& R@3\,$\uparrow$
& MM-Dist\,$\downarrow$
& Div.\,$\rightarrow$
& Penet.\,$\downarrow$
& Float\,$\downarrow$
& Skate\,$\downarrow$
& Phys-Err\,$\downarrow$
& Jerk\,$\downarrow$
& NDMS\,$\uparrow$ \\
\midrule

\rowcolor{gray!20}
\multicolumn{13}{c}{\bf Train split (3485 sequences)} \\

\textit{Mocap (reference)}
& --
& --
& \textit{0.669}
& \textit{0.963}
& \textit{1.60}
& \textit{15.76}
& \textit{15.62}
& \textit{6.56}
& \textit{7.45}
& \textit{29.27}
& \textit{38.9}
& \textit{0.579} \\

1-shot kin. planner
& 42.81
& 96.5
& 0.251
& 0.491
& 7.39
& 13.77
& 0.02
& 14.08
& 9.55
& 23.65
& 478.2
& 0.375 \\

1-shot kin. planner + FT
& 76.50
& 125.2
& 0.269
& 0.536
& 6.51
& 13.81
& 0.02
& 16.05
& 10.94
& 27.01
& 655.1
& 0.359 \\

Replan. kin. planner
& 15.95
& 70.7
& 0.159
& 0.365
& 9.24
& 13.33
& 0.06
& 18.00
& 10.02
& 28.08
& 567.2
& 0.340 \\

Replan. kin. planner + FT
& 33.97
& 157.9
& 0.136
& 0.326
& 9.75
& 11.14
& 0.05
& 16.84
& 9.59
& 26.48
& 653.1
& 0.297 \\

\textbf{Ours (LYRIC)}
& \textbf{91.33}
& 135.2
& \textbf{0.374}
& \textbf{0.698}
& \textbf{4.61}
& \textbf{15.08}
& \textbf{0.02}
& \textbf{13.25}
& \textbf{7.19}
& \textbf{20.46}
& \textbf{281.7}
& \textbf{0.452} \\

\midrule

\rowcolor{gray!20}
\multicolumn{13}{c}{\bf Held-out split (267 sequences)} \\

\textit{Mocap (reference)}
& --
& --
& \textit{0.488}
& \textit{0.853}
& \textit{2.91}
& --
& \textit{12.76}
& \textit{8.95}
& \textit{8.06}
& \textit{29.77}
& \textit{41.0}
& \textit{0.570} \\

1-shot kin. planner
& 37.08
& 88.8
& 0.162
& 0.368
& 9.48
& --
& \textbf{0.00}
& 8.82
& 9.19
& 18.01
& 475.7
& 0.344 \\

1-shot kin. planner + FT
& 74.16
& 120.1
& 0.200
& 0.429
& 8.10
& --
& 0.04
& 13.18
& 10.89
& 24.11
& 647.4
& 0.334 \\

Replan. kin. planner
& 10.49
& 60.6
& 0.127
& 0.314
& 10.26
& --
& \textbf{0.00}
& 12.54
& 10.14
& 22.68
& 563.0
& 0.335 \\

Replan. kin. planner + FT
& 37.83
& 108.5
& 0.107
& 0.277
& 10.45
& --
& \textbf{0.00}
& 13.33
& 10.13
& 23.45
& 671.0
& 0.290 \\

\textbf{Ours (LYRIC)}
& \textbf{90.26}
& 148.8
& \textbf{0.361}
& \textbf{0.666}
& \textbf{4.94}
& --
& 0.02
& \textbf{7.08}
& \textbf{6.75}
& \textbf{13.85}
& \textbf{256.7}
& \textbf{0.451} \\

\bottomrule
\end{tabular}}
\end{table*}

\paragraph{Effectiveness of planner-guided post-tuning.}
We next evaluate Stage~III, where the frozen planner provides object–root trajectory targets for on-policy post-tuning of the action generator.
On-policy post-tuning (\emph{Flow-BC + FT}) improves success from $80.32\%$ to $91.33\%$ on the training split and from $74.16\%$ to $90.26\%$ on the held-out subject.
These gains demonstrate that post-tuning improves task execution and that this improvement generalizes to the held-out subject.

On the training split, post-tuning introduces small trade-offs in retrieval accuracy, with R@1 decreasing from $0.390$ to $0.374$ and R@3 decreasing from $0.709$ to $0.698$.
However, MM-Dist remains nearly unchanged, while Diversity moves closer to the reference value.
On the held-out split, all reported semantic metrics improve: R@1 increases from $0.347$ to $0.361$, R@3 from $0.626$ to $0.666$, and MM-Dist decreases from $5.56$ to $4.94$.
Thus, although the planner-guided post-tuning stage primarily targets reliable physical execution, it preserves the language-conditioned behavior learned during offline training and improves semantic generalization to the held-out subject.

Post-tuning also improves nearly all physical and temporal motion-quality metrics on the held-out split.
Float decreases from $8.12$ to $7.08$\,mm, Skate from $7.00$ to $6.75$\,mm, Phys-Err from $15.12$ to $13.85$\,mm, and Jerk from $281.2$ to $256.7\,\mathrm{m}/\mathrm{s}^{3}$, while NDMS increases from $0.440$ to $0.451$.
Penetration remains negligible for both models.
On the training split, post-tuning similarly reduces Skate and Jerk and improves NDMS, although Float and Phys-Err increase slightly.
The substantial success gains are therefore not accompanied by a broad degradation in motion quality.
Instead, post-tuning improves temporal quality and most physical metrics, particularly on the held-out subject.
The remaining gap to the Mocap reference in Jerk and NDMS may reflect additional temporal variation relative to kinematic motion capture introduced by physics-based control, including contact forces, feedback corrections, and high-frequency actuation.

\paragraph{Qualitative results.}
Figure~\ref{fig:bcvsrl} compares Flow-BC and Flow-BC + FT on representative contact-rich interactions.
Flow-BC generally captures the intended behavior, but small execution errors can compound into unstable whole-body compensation or loss of object control, as seen during overhead chair transport, box rotation, and floor-lamp relocation.
After post-tuning, the controller maintains better balance and more stable object contact through these challenging phases, allowing execution to continue toward the requested placement.
These examples complement the quantitative gains in Table~\ref{tab:flow_policy} by illustrating the improved physical execution achieved by planner-guided post-tuning, where the frozen planner provides task-level trajectory supervision during on-policy refinement.

Figure~\ref{fig:gallery} shows representative rollouts of our final controller across objects with substantially different geometry and scale.
For the same object, different instructions produce distinct interaction strategies: the controller lifts, pulls, or kicks the floor lamp; pulls, carries, or kicks the plastic box; carries the table normally or overhead; and lifts, pulls, or pushes the suitcase.
These behaviors require different contact choices and coordinated whole-body motion, demonstrating that language controls the interaction semantics and manner rather than merely specifying an object displacement.
We provide additional video results for our final generative controller, as well as a comparison between Flow-BC and Flow-BC + FT, on the supplementary webpage.

\subsection{Kinematic-Planner Baselines}
\label{sec:exp_baselines}

\paragraph{Baseline construction.}
An alternative to our factorized controller is to use a text-conditioned kinematic motion generator as the planner and execute its predicted dense full-body humanoid and object trajectories with a tracking policy, as in prior language-driven physics systems centered on full-body human-motion generation~\citep{wu2025human,tevet2024closd,lin2025simgenhoi}.
We construct two baselines following this approach. 
In both, a text-conditioned rectified-flow generator predicts per-frame humanoid root poses, joint rotations, and object poses. 
Our unified tracking policy from Sec.~\ref{sec:tracking} uses the current simulated state to track these predicted trajectories.
The two generators share the same architecture, motion representation, and text encoder, but are trained separately for one-shot generation and short-horizon replanning, which we elaborate below.
The first baseline uses one-shot kinematic planning followed by tracking. 
At each episode reset, it generates a whole-sequence kinematic reference conditioned on the text instruction, initial character--object state, and target object pose, and we track this entire reference without replanning. 
Following prior diffusion-based motion generation methods~\citep{tevet2022human}, we cap its length to $T=\min(T_{\mathrm{demo}},300)$, where $T_{\mathrm{demo}}$ is the duration of the corresponding ground-truth mocap sequence.
If the episode continues beyond the reference, the tracker continues following its final frame.
The other baseline follows CLoSD-style closed-loop replanning~\citep{tevet2024closd}: it conditions on recent 30-frame simulated motion, predicts a 60-frame kinematic horizon, executes 32 frames using our tracker, and then replans.
To compare the planning interfaces under analogous post-tuning, the fine-tuned variants keep each kinematic planner frozen and update only its tracker, mirroring our frozen-planner post-tuning of the action generator.
Tracker fine-tuning combines the tracking reward from Sec.~\ref{sec:tracking} against the generated kinematic reference, with the task-completion bonus $b_{\mathrm{goal}}\mathbf{1}_{\mathrm{goal}}(t)$ from Sec.~\ref{sec:flow_rl}.
The combined generator--tracker parameter budget is matched to our controller, and all methods use the same evaluation protocol.

\paragraph{Results.}

Table~\ref{tab:flow_baseline_comparison} shows that tracker fine-tuning
(+FT) substantially improves task success for both kinematic-planner
baselines.
These gains demonstrate the benefit of adapting the tracker to generated
references and the task-completion reward, even while keeping the
kinematic planner frozen.

Nevertheless, the strongest kinematic-planner baseline (1-shot kin. planner + FT) reaches only $76.50\%$ success on the training split and $74.16\%$ on the held-out split, compared with $91.33\%$ and $90.26\%$ for our direct action controller.
The replanning baseline performs substantially worse than its one-shot counterpart both before and after tracker
fine-tuning.
Although the one-shot baseline can not update its kinematic reference during execution, it generates the complete
motion from a mocap-like initial condition and retains a single coherent
kinematic trajectory.
The replanning baseline instead repeatedly conditions on simulator-realized histories that
contain accumulated tracking errors, potentially moving the conditioning states away from the kinematic generator's training distribution.
This distribution shift can compound across replanning cycles, causing
the generated reference to progressively drift in both task intent and
motion coherence, consistent with that baseline's substantially weaker semantic
metrics.
This effect is particularly severe in contact-rich object interaction, where the object is unactuated, so small execution errors can therefore alter both the object state and the subsequent human--object interaction context seen by the kinematic generator.
Fine-tuning the tracker improves its ability to execute these references
but cannot correct drift produced by the frozen generator.
Thus, the comparison does not suggest that closed-loop feedback is
inherently harmful; rather, it shows that replanning is ineffective when
the generator does not generalize to the physical-state distribution
created by its own execution.

In contrast, our controller learns action distributions from physically executed rollouts and updates its actions using the latest simulated state and contacts.
Its planner specifies short-horizon object and humanoid-root trajectories rather than a dense full-body reference, leaving the action policy free to adjust joint motions and contact configurations during execution.
During post-tuning, the same object–root trajectories also provide task-level supervision for refining physical execution.

Our method also provides stronger overall semantic and physical
quality than the kinematic-planner baselines, with better R-Precision
and MM-Dist, lower physical error and Jerk, and higher NDMS.
Notably, although tracker fine-tuning increases the success of both
baselines, it also increases Jerk and reduces NDMS.
Fine-tuning improves the tracker, but the pipeline still requires it
to realize a separately generated full-body kinematic reference under
physical dynamics and contact constraints, leaving a potential mismatch
between motion generation and execution.
Our controller avoids this full-body generator--tracker interface by
directly generating actions from the current simulated state and contacts,
with further adaptation through on-policy post-tuning.
This design achieves higher task success together with better semantic
and motion quality than the kinematic-planner baselines.
Additional qualitative comparisons with the kinematic-planner baselines are provided on the supplementary webpage.

\begin{figure}[!t]
    \centering
    \includegraphics[width=0.45\textwidth]{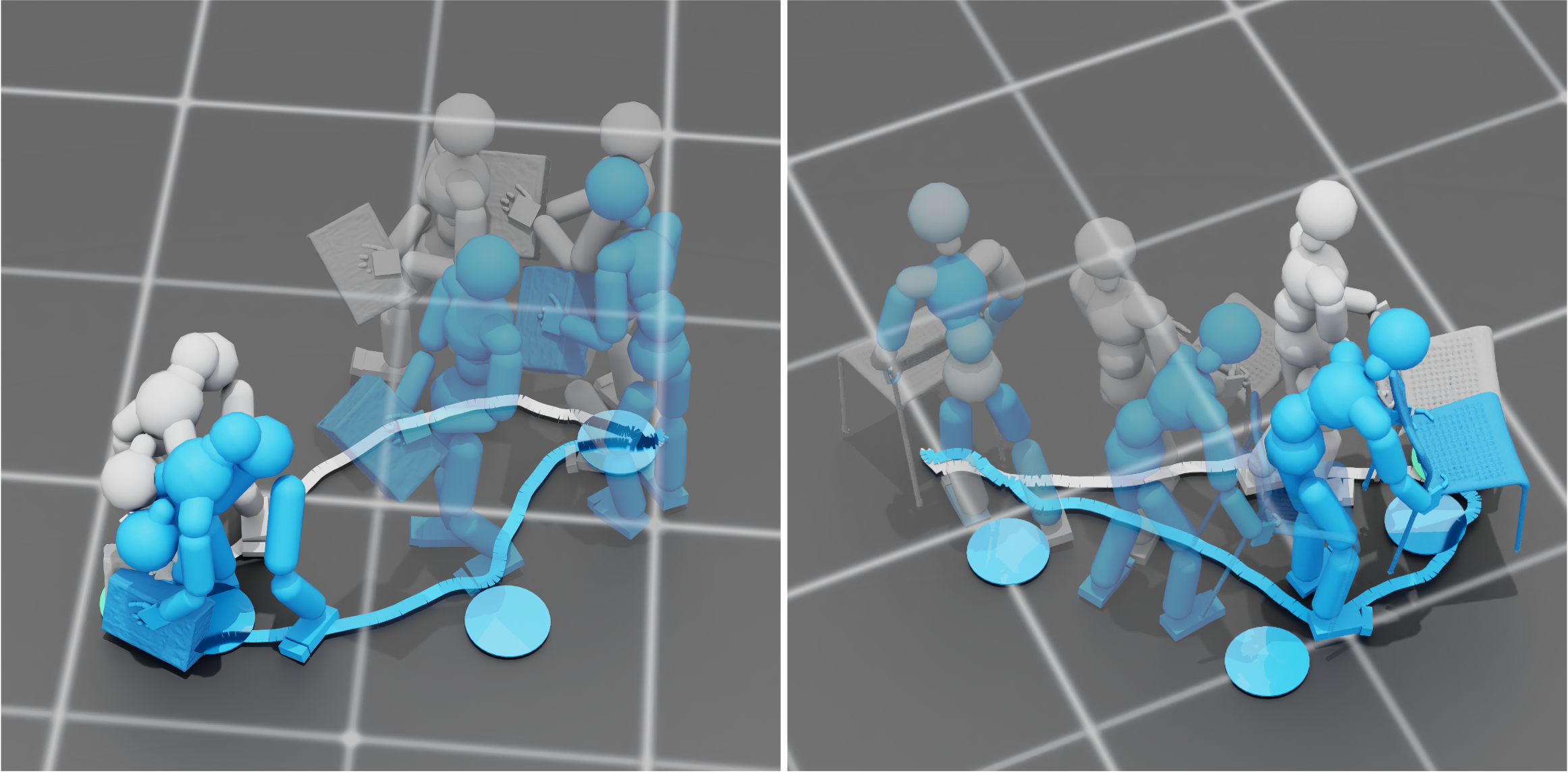}
        \caption{\textbf{Representative object waypoint-guided interactions.}
        \textcolor{blue!70}{Blue disks} indicate user-specified object waypoints; \textcolor{blue!70}{blue} and \textcolor{gray}{gray} curves show the guided and unguided simulator-realized object trajectories, respectively.
        Guidance steers the object toward the intermediate waypoints.}
    \label{fig:waypoints}
\end{figure}

\subsection{Test-Time Spatial Control}
\label{sec:exp_waypoint}

We further test whether the learned trajectory planner can support user-specified spatial constraints at test time without retraining.
Following CHOIS~\citep{li2024controllable}, we specify $k\in\{1,3,7\}$ intermediate object waypoints for each sequence in the held-out split.
We constrain only the horizontal $(x,y)$ object position, following the CHOIS waypoint-control setting, so that the user specifies where the object should move while leaving its height and orientation free to adapt to the interaction.
For this experiment, both guided and unguided variants use 20 planner denoising steps, while the action sampler remains unchanged, so that guidance is the only varying factor.

\paragraph{Waypoint guidance mechanism.}
We adopt gradient-based test-time guidance, following the general principle of guided generative sampling~\citep{dhariwal2021diffusion,chung2022diffusion,bansal2024universal}.
Our factorized design allows us to apply spatial guidance to the trajectory planner while the action generator adapts whole-body actions to realize the guided plan.
The planner predicts only a 30-frame (1\,s) horizon, whereas a waypoint may lie several seconds into the future.
We therefore use two guidance regimes.
At each planner update, we determine whether the waypoint's time frame lies within the current planning horizon.
If it does, we apply full-strength guidance to the object-position token at the corresponding plan time frame.
Otherwise, we apply weaker, temporally decayed guidance to the final object-position token, encouraging progress toward the waypoint before its specified time frame is covered by the plan.
As successive updates advance the planning horizon, guidance therefore transitions from influencing the plan endpoint to enforcing the waypoint at its specified time. See appendix for more detailed explanation.

\begin{table}[t]
\centering
\caption{\textbf{Test-time object waypoint control on the held-out OMOMO split.}
Waypoint errors are measured in centimeters at the commanded frames; Plan and Exec. denote the predicted plan and physically executed object trajectory, respectively.
}
\label{tab:waypoint_control}
\setlength{\tabcolsep}{5pt}
\begin{tabular}{l cc cc cc}
\toprule
& \multicolumn{2}{c}{$k=1$}
& \multicolumn{2}{c}{$k=3$}
& \multicolumn{2}{c}{$k=7$} \\
\cmidrule(lr){2-3}
\cmidrule(lr){4-5}
\cmidrule(lr){6-7}
Method
& Plan & Exec.
& Plan & Exec.
& Plan & Exec. \\
\midrule
No guidance
& 22.35 & 38.79
& 21.80 & 30.73
& 21.18 & 29.79 \\
+ Guidance
& \textbf{10.43} & \textbf{20.48}
& \textbf{10.61} & \textbf{17.68}
& \textbf{12.07} & \textbf{14.51} \\
\bottomrule
\end{tabular}
\end{table}

\paragraph{Results.}
We evaluate waypoint accuracy using horizontal Euclidean position errors, measured in centimeters at the commanded frames.
Plan error measures the distance between each waypoint and the object position predicted by the planner for its commanded frame; Exec. error measures the corresponding deviation of the physically simulated object.
As shown in Table~\ref{tab:waypoint_control}, guidance consistently improves waypoint accuracy across all three control settings.
For $k=1$, $3$, and $7$, Plan error decrease from $22.35$, $21.80$, and $21.18$\,cm to $10.43$, $10.61$, and $12.07$\,cm, respectively.
The improvement remains substantial after physical execution, with Exec. error decreasing from $38.79$, $30.73$, and $29.79$\,cm to $20.48$, $17.68$, and $14.51$\,cm.
These results show that the test-time guidance effectively steers the learned planner toward user-specified spatial constraints, and that this control transfers to the physically simulated object.
The larger Exec. error relative to Plan error reflects the additional difficulty of realizing the guided trajectory through contact-rich whole-body control.
Figure~\ref{fig:waypoints} visualizes representative waypoint-guided interactions.
The guided object trajectories are steered toward the specified intermediate waypoints while the controller maintains physically plausible whole-body interaction.
More video results of waypoint control are provided on the supplementary webpage.

Spatial guidance introduces a moderate trade-off in terminal task completion.
Relative to $89.89\%$ success without guidance, placement success becomes $85.39\%$, $86.14\%$, and $83.15\%$ for $k=1$, $3$, and $7$, respectively.
Guidance imposes additional spatial and temporal constraints on the object trajectory while keeping the action policy fixed.
The resulting plans can be more difficult to execute in that case, causing some episodes to terminate due to plan divergence before final placement.
Thus, the proposed guidance substantially improves intermediate object-trajectory control while largely preserving the original terminal placement capability, without any additional training.

\begin{figure}[!t]
    \centering
    \includegraphics[width=0.45\textwidth]{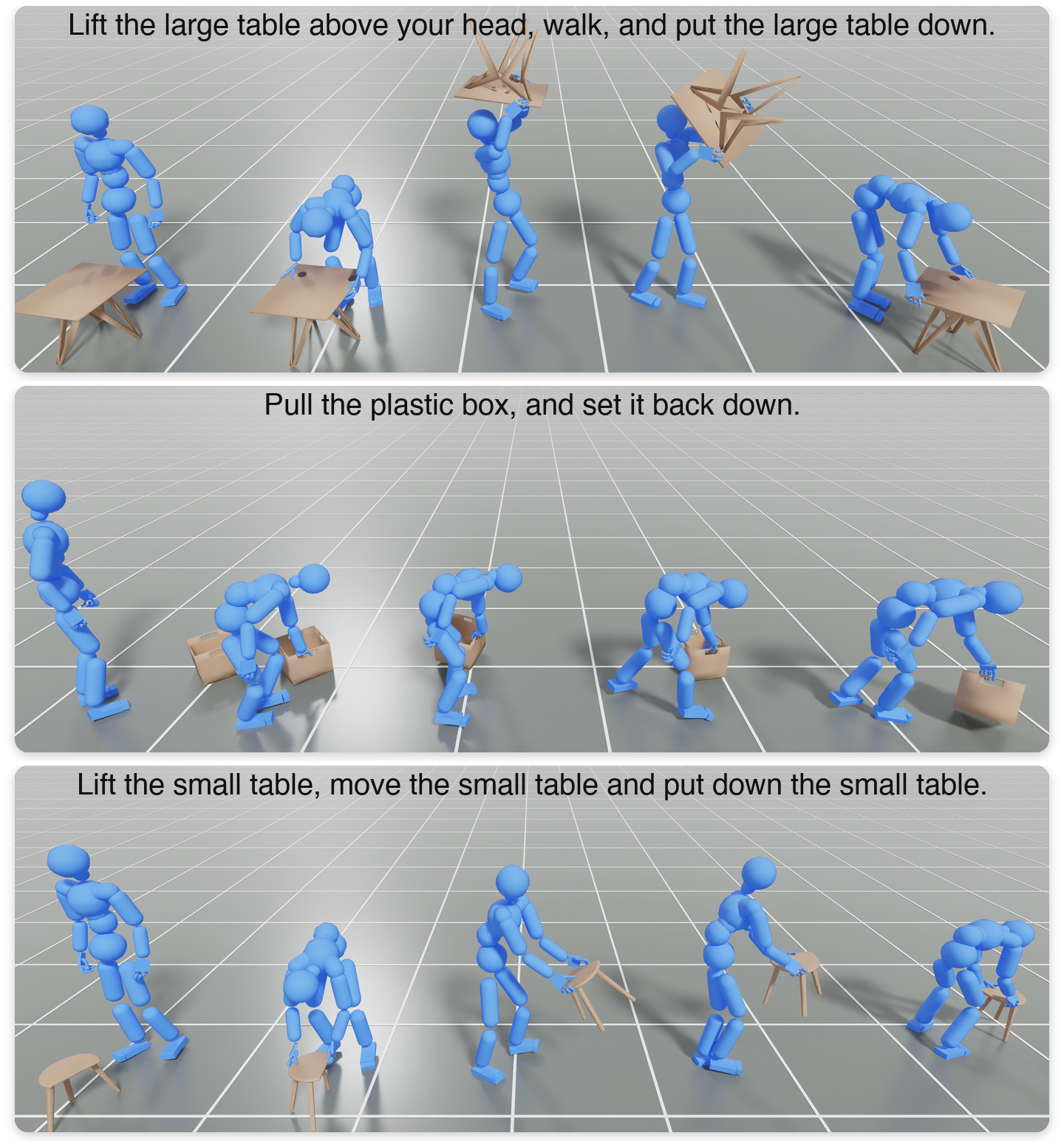}
        \caption{\textbf{Zero-shot generalization to novel object shapes.}
        Our policy successfully performs diverse interactions with novel within-category object geometries without fine-tuning.
        }
    \label{fig:new_object_shape}
\end{figure}

\subsection{Zero-Shot Generalization to Novel Object Shapes}
To qualitatively evaluate zero-shot generalization to novel object geometry, we replace each test object with several semantically matched meshes from Objaverse~\citep{deitke2023objaverse}, while retaining the original text prompt.
Each mesh is orientation-aligned, uniformly scaled, and center-registered without policy fine-tuning. 
As shown in Fig.~\ref{fig:new_object_shape}, the frozen policy transfers across substantial within-category shape variations while maintaining coordinated whole-body motion and object control, including lifting, carrying, pulling, and fine-grained fingertip grasps.
More results are provided in the appendix and supplementary webpage.

\begin{table}[t]
\centering
\caption{\textbf{User study on motion tracking and language-driven character control.}
Each cell is the percentage of trials in which raters chose that method.
\emph{Fidelity} asks which rollout best follows the mocap reference (tracking)
or the written instruction (language-driven control);
\emph{Naturalness} asks which moves most naturally.}
\label{tab:user-study}
\begin{tabular}{@{}lcc@{}}
\toprule
& \multicolumn{2}{c}{Preference (\%) $\uparrow$} \\
\cmidrule(l){2-3}
Method & Fidelity & Naturalness \\
\midrule
\rowcolor{gray!25} \multicolumn{3}{c}{Interaction tracking policy (2 alternatives, chance 50\%)} \\
InterMimic~\cite{xu2025intermimic} & 35.3 & 21.2 \\
\textbf{Ours} & \textbf{64.7} & \textbf{78.8} \\
\rowcolor{gray!25} \multicolumn{3}{c}{Language-driven control (4 alternatives, chance 25\%)} \\
1-shot kin. planner + FT & 1.7 & 1.1 \\
Replan. kin. planner + FT & 0.8 & 0.4 \\
Flow-BC & 30.1 & 25.2 \\
\textbf{Flow-BC + FT} & \textbf{67.4} & \textbf{73.3} \\
\bottomrule
\end{tabular}
\end{table}

\subsection{User Study}

We conduct a blinded user study to evaluate the perceptual quality of both stages.
For tracking, raters compare our subject-specific tracking policy with InterMimic given the mocap reference.
Given the language instruction, raters compare our behavior-cloned controller, its on-policy post-tuned version, and the one-shot and replanning kinematic-planner baselines with fine-tuned trackers.
We evaluate both fidelity to the conditioning signal and motion naturalness, with full study details provided in the appendix.

Table~\ref{tab:user-study} shows a clear preference for our methods.
For tracking, our tracker is preferred over InterMimic in both fidelity ($64.7\%$) and naturalness ($78.8\%$), supporting the perceptual benefit of the geometry-conditioned hand-interaction costs.
For language-driven control, our on-policy post-tuned controller receives $67.4\%$ of the fidelity preference and $73.3\%$ of the naturalness preference, substantially outperforming its behavior-cloned counterpart and both kinematic-planner baselines with fine-tuned trackers.
These results support the perceptual benefits of on-policy post-tuning for instruction adherence and motion naturalness, complementing the improvements in task success.

\section{Conclusion}
We presented LYRIC, a generative flow-matching controller for contact-rich whole-body object interaction from a free-form language instruction and a sparse terminal object goal.
Geometry-conditioned interaction rewards and relaxed reference tracking allow a unified tracking policy to convert imperfect motion-capture references into physically executed expert trajectories with reliable hand interaction.
The controller separates task progression from motor execution: a task-level planner predicts short-horizon object and humanoid-root trajectories, while an action generator resolves whole-body motion and contacts in closed loop.
During on-policy post-tuning, the frozen planner provides stable supervision for intermediate task progression as the action generator learns from its own rollouts.
At deployment, the controller requires neither the tracking policy nor a prescribed dense full-body motion.
Controlled evaluations show higher tracking success than an InterMimic reimplementation and higher task success, semantic alignment, and motion quality than matched kinematic-planner baselines.
The factorization also supports test-time object-waypoint guidance and qualitative transfer to novel within-category shapes.
Future work could extend the framework to articulated-object and multi-object interactions and study generalization across broader geometric and semantic variation.

\bibliographystyle{ACM-Reference-Format}
\bibliography{paper}

\clearpage
\appendix

\appendix

\noindent\textbf{Appendix contents.}
\begin{enumerate}
    \item \hyperref[app:tracking]{Tracking policy details}
    \item \hyperref[app:flow]{Text-conditioned controller}
    \item \hyperref[app:post_tuning]{On-policy post-tuning}
    \item \hyperref[app:evaluation]{Evaluation details}
    \item \hyperref[app:human_study]{Human study}
\end{enumerate}

\section{Tracking Policy Details}
\label{app:tracking}

\subsection{Simulation and Tracking Setup}
\label{app:simulation}

\begin{table*}[t]
\centering
\small
\setlength{\tabcolsep}{3pt}
\caption{\textbf{Physical settings and unified-tracker training parameters.} PD gains are stiffness/damping; friction values apply to both static and dynamic friction. }
\label{tab:app_physics}
\label{tab:app_tracking_ppo}
\begin{tabular}{@{}p{0.27\textwidth}p{0.18\textwidth}p{0.27\textwidth}p{0.20\textwidth}@{}}
\toprule
Physics and control & Value & PPO & Value \\
\midrule
Reference lookahead & $1$ and $16$ steps & Parallel environments & $16{,}384$ \\
Position / velocity iterations & $4/1$ & Rollout length & $32$ steps \\
Self-collision / continuous detection & Off / on & Actor / critic minibatch & $131{,}072/65{,}536$ \\
Object density & $200$\,kg/$\mathrm{m}^3$ & Update epochs per rollout & $5$ \\
Ground / object friction & $0.9/0.6$ & Optimizer & Adam \\
Ground / object restitution & $0.1/0.05$ & Initial actor / critic learning rate & $3\times10^{-5}$ each \\
Object contact / rest offset & $0.02/0$\,m & Actor / critic schedule & Linear / constant \\
Object linear / angular damping & $0.01/0.01$ & Discount / GAE parameter & $0.99/0.95$ \\
Leg PD gains & $800/80$ & PPO clipping & $0.2$ \\
Torso PD gains & $1000/100$ & Actor gradient-norm limit & $1.0$ \\
Arm, neck, and head PD gains & $500/50$ & Fixed exploration std. & $0.055$ \\
Finger PD gains & $100/10$ & Entropy coefficient & $0$ \\
Actuator effort limit & $3000$\,N\,m & Normalized observation clipping & $[-5,5]$ \\
Actuator velocity limit & $50$\,rad/s & Action clipping & $[-1,1]$ \\
& & Teacher checkpoint & Epoch $34{,}000$ \\
\bottomrule
\end{tabular}
\end{table*}

\begin{table}[t]
\centering
\small
\caption{\textbf{Actor observation perturbations.} Each entry is the half-width of an independent component-wise uniform distribution, applied before observation normalization.}
\label{tab:app_noise}
\begin{tabular}{@{}p{0.46\columnwidth}p{0.22\columnwidth}p{0.22\columnwidth}@{}}
\toprule
Feature & Humanoid & Object \\
\midrule
Root height & $0.02$\,m & --- \\
Position & $0.05$\,m & $0.07$\,m \\
6D rotation components & $0.05$ & $0.07$ \\
Linear velocity & $0.35$\,m/s & $0.50$\,m/s \\
Angular velocity & $0.25$\,rad/s & $0.35$\,rad/s \\
Surface-proximity vectors & $0.03$ & --- \\
\bottomrule
\end{tabular}
\end{table}

The supplied InterMimic colliders approximate thin structures such as poles too coarsely, leaving a mismatch between the collision surface and visible geometry.
A hand can consequently make simulated contact while appearing displaced from the object.
We regenerate the colliders with CoACD~\citep{wei2022approximate}, using a concavity threshold of $0.03$, at most 16 convex pieces, and at most 64 vertices per piece.
Both our tracker and the InterMimic policies used for comparison are trained with these same regenerated colliders.

For canonical-body retargeting, we adapt OmniRetarget~\citep{yang2025omniretarget} to optimize root and joint poses between SMPL-X bodies with the same topology.
The optimization preserves object-relative contact locations and stance feet while penalizing penetration and abrupt pose changes.
The object trajectory stays fixed during optimization; afterward, we translate the humanoid and object together to align the sequence with the ground.

For the tracker, interaction feature $\boldsymbol{\phi}_t$ follows InterMimic~\citep{xu2025intermimic}: joint-to-nearest-surface proximity vectors for the body and fingers, and simulated contact indicators at non-root joints.
The proximity vectors encode surface direction and proximity in the humanoid heading frame.

We also report the hyperparameters in the following tables:
Table~\ref{tab:app_physics} lists the physical settings and tracking-policy training parameters, and Table~\ref{tab:app_noise} specifies the observation-noise ranges.
Table~\ref{tab:app_interaction_weights} lists the tracking reward coefficients.

\begin{table}[t]
\centering
\small
\caption{\textbf{Tracking reward coefficients}. }
\label{tab:app_interaction_weights}
\begin{tabular}{@{}p{0.62\columnwidth}p{0.32\columnwidth}@{}}
\toprule
Term & Coefficient \\
\midrule
Humanoid position / rotation & $30/1.5$ \\
Object position / rotation / linear velocity & $5/0.1/0.1$ \\
Interaction geometry & $5$ \\
Required / undesired contact & $5/3$ \\
Hand-contact sharpness & $5$ \\
\midrule
Geometry-conditioned grasp & $0.15$ \\
Palm alignment & $0.35$ \\
Reference contact anchor & $1.5$ \\
Hold stability & $0.05$ \\
Joint / object acceleration & $10^{-4}/(2\times10^{-5})$ \\
Contact force & $10^{-9}$ \\
\bottomrule
\end{tabular}
\end{table}

\subsection{Geometry-Conditioned Grasp Rewards}
\label{app:grasp}

\paragraph{Local thickness.}
We estimate surface thickness from interior tangent-sphere diameters~\citep{inui2016shrinking}.
Near $\mathbf r$, a Gaussian-weighted average with spatial scale $2$\,cm smooths variations between surface samples.
Writing this local average as $\operatorname{thick}(\mathbf r)$, the mode weight is
\begin{equation}
\alpha=\operatorname{sigmoid}\!\left(
\frac{0.08\,\mathrm m-\operatorname{thick}(\mathbf r)}{0.02\,\mathrm m}
\right).
\label{eq:app_alpha}
\end{equation}
Thus $8$\,cm is the transition thickness, i.e., at $8$\,cm, enclosure mode and support mode get the same weights, and $2$\,cm controls its softness.

\paragraph{Cross-section construction.}
To construct the cross section, we estimate the local structure's long axis and interior center.
We obtain $\mathbf{k}$ by applying weighted PCA to nearby surface normals and selecting the direction of smallest variance.
We then move nearby surface points inward along their normals by half the local thickness.
For a cylindrical pole, this corresponds to moving inward by approximately one radius toward its centerline.
We average the resulting interior points using the same neighborhood weights and place the cross-sectional plane through this estimated center, perpendicular to $\mathbf{k}$.

\paragraph{Sector occupancy $o_b$.}
Let $\chi_i\in\{0,1\}$ indicate whether finger joint $i$ contacts the object.
We project the contacting joints onto the cross-sectional plane and divide their directions around its center into $B=16$ angular sectors.
Each contacting joint contributes a total of one, distributed mainly to the sector containing it and partly to neighboring sectors to avoid abrupt changes at sector boundaries.
For each sector $b$, we sum these contributions from all contacting joints and denote the result by $num_b$.
The occupancy score is then
\begin{equation}
    o_b=1-\exp(-5num_b).
\label{eq:app_occupancy}
\end{equation}
The score increases with accumulated contact but saturates, so additional contacts in an already occupied direction provide diminishing benefit.
For $\bar o_b$, we sum the occupancies on the opposite side of the cross section and apply the same saturation with coefficient $3$.

\paragraph{Thumb and finger gates.}
We count the contacting joints on the thumb and on the remaining four fingers, denoting these counts by $num_{\mathrm{thumb}}$ and $num_{\mathrm{other}}$, respectively.
We also count how many of the four non-thumb fingers have at least one joint in contact and denote this number by $num_{\mathrm{finger}}$.
The gates are
\begin{equation}
\begin{aligned}
G^{\mathrm{tf}}
    &=\bigl(1-e^{-num_{\mathrm{thumb}}}\bigr)
      \bigl(1-e^{-num_{\mathrm{other}}}\bigr),\\
G^{\mathrm{part}}
    &=\left(\frac{num_{\mathrm{finger}}}{4}\right)^{0.75}.
\end{aligned}
\label{eq:app_finger_gates}
\end{equation}
The first gate requires contact from both the thumb and the other fingers: it is zero if either group has no contact.
The second gate increases with the number of participating fingers, so several contacting joints on one finger cannot replace contact from multiple fingers.

\paragraph{Support gate and spread.}
We average the vectors from nearby object-surface points to the palm to estimate the outward surface direction.
We then project all 15 finger joints onto the plane perpendicular to this direction.
Their covariance $\boldsymbol{\Sigma}_{\mathrm{tan}}$ is computed with equal weights, dividing the sum by 15.
The quantity $A=\sqrt{\det\boldsymbol{\Sigma}_{\mathrm{tan}}}$ measures how widely the joints spread across the plane.
It is small when the joints cluster together or lie nearly along a line, and grows when they spread in both directions.

We use the total number of contacting joints, $num_{\mathrm{thumb}}+num_{\mathrm{other}}$, to compute the contact gate, and normalize the spread measure through $\psi$:
\begin{equation}
\begin{aligned}
G^{\mathrm{contact}}
    &=1-\exp\!\left[-0.3\bigl(num_{\mathrm{thumb}}+num_{\mathrm{other}}\bigr)\right],\\
\psi(A)&=
\begin{cases}
0, & A\leq A_0,\\
\left(\dfrac{A-A_0}{A_1-A_0}\right)^{1.8}, & A_0<A<A_1,\\
1, & A\geq A_1.
\end{cases}
\end{aligned}
\label{eq:app_support}
\end{equation}
The contact gate is zero without contact and approaches one as more joints make contact.
For the spread term, $A_0=3\times10^{-4}\,\mathrm m^2$ is the lower threshold below which no spread credit is given, while $A_1=10^{-3}\,\mathrm m^2$ is the threshold for full credit.

\section{Text-Conditioned Controller}
\label{app:flow}

\subsection{Interaction Features}
\label{app:flow_features}
Table~\ref{tab:flow_features} specifies the action generator's interaction features.
The contact indicators are the same as the tracker's; signed distances are positive outside the object, and their unit gradients give the outward surface direction.
\begin{table}[t]
\centering
\small
\caption{\textbf{Action-generator interaction features $\boldsymbol{\phi}_t$.} Direction vectors use the humanoid heading frame.}
\label{tab:flow_features}
\begin{tabular}{@{}p{0.62\columnwidth}p{0.32\columnwidth}@{}}
\toprule
Feature & Dimensions \\
\midrule
Scene-contact indicators, excluding pelvis & $51$ \\
Object signed distance at each body joint & $52$ \\
Unit signed-distance gradient & $52\times3$ \\
Local thickness at the nearest surface point & $52$ \\
Local linearity and planarity & $52\times2$ \\
Patch descriptors at two palms and pelvis & $3\times7$ \\
\midrule
Total & $436$ \\
\bottomrule
\end{tabular}
\end{table}

\paragraph{Local shape descriptors.}
For the shape features, we use Gaussian-weighted neighborhoods of surface points with spatial scale $5$\,cm.
Linearity and planarity measure whether the neighboring surface points lie mainly along a line or within a plane.

Each patch descriptor is evaluated at the surface point nearest to the corresponding palm or pelvis.
A palm query uses the mean position of its five proximal finger joints.
The seven values are the dominant elongation direction (three), the extent along and perpendicular to it (two), and the local surface centroid's offset from the query surface point in the perpendicular plane (two).
The two extents are twice the standard deviation along the dominant direction and the root-mean-square distance from that axis, respectively.

\subsection{Training Settings}
\label{app:flow_training}
\paragraph{Text augmentation.}
To improve robustness to different phrasings of the same interaction, for each original instruction, we prepare eight alternative phrasings that preserve the same interaction and manner of execution.
During training, we randomly select the original instruction or one of these paraphrases, allowing the model to encounter different wording for the same motion.
We encode the selected text using CLIP ViT-B/32, which produces a $512$-dimensional sentence embedding.
Text conditioning is dropped with probability $0.1$ during training; classifier-free guidance is disabled at deployment.

\paragraph{Plan conditioning.}
The probability of conditioning on a sampled plan instead of training data increases linearly from zero at update $600{,}000$ to one at $900{,}000$.
For sampled plans, the number of Euler steps is sampled uniformly from $\{2,3,4,5\}$.

\begin{table}[t]
\centering
\small
\caption{\textbf{Additional settings for Stage~II.}}
\label{tab:flow_training}
\begin{tabular}{@{}p{0.62\columnwidth}p{0.32\columnwidth}@{}}
\toprule
Setting & Value \\
\midrule
AdamW weight decay & $10^{-4}$ \\
Gradient-norm clipping & $1$ \\
Warmup / start of cosine decay & $10^4/9\times10^5$ updates \\
EMA decay & $0.9999$ \\
Planner heads / feed-forward width & $4/1024$ \\
Action heads / feed-forward width & $8/2048$ \\
\bottomrule
\end{tabular}
\end{table}

\subsection{Waypoint Guidance}
\label{app:waypoints}
At each flow step, we estimate the final trajectory and adjust the current sample to bring its predicted object positions closer to the requested waypoints.
Let $\mathbf{v}_{\mathrm P}$ denote the planner's velocity field, with conditioning omitted for brevity.
From the current trajectory sample $\mathbf{x}_{\tau}$, we estimate the final trajectory in normalized coordinates as
$\hat{\mathbf{P}}=\mathbf{x}_{\tau}+(1-\tau)\mathbf{v}_{\mathrm P}(\mathbf{x}_{\tau},\tau)$.
We convert the estimated object positions to world coordinates, denoting the position at frame $f$ by $\mathbf{p}^{\mathrm{obj}}_f(\hat{\mathbf{P}})$.
For a waypoint target $\mathbf{w}_f$, we measure the squared horizontal distance between the predicted and requested positions:
\begin{equation}
\mathcal{J}(\hat{\mathbf{P}})
=\frac12\sum_f\omega_f
\left\|
\left[\mathbf{p}^{\mathrm{obj}}_f(\hat{\mathbf{P}})-\mathbf{w}_f\right]_{xy}
\right\|_2^2.
\label{eq:app_waypoint_loss}
\end{equation}
The weights $\omega_f$ determine which predicted positions receive guidance.
A waypoint within the planning horizon receives weight one at its requested frame.
When no waypoint lies within the horizon, we instead guide the final predicted position toward the next waypoint, using weight $0.15$ times the planning horizon divided by the time remaining until that waypoint.
Both durations are measured in control steps, and all other positions receive zero weight.
We then combine the usual flow update with a gradient step that reduces the waypoint error:
\begin{equation}
\mathbf{x}_{\tau+\Delta\tau}
=\mathbf{x}_{\tau}
+\Delta\tau\,\mathbf{v}_{\mathrm P}(\mathbf{x}_{\tau},\tau)
-\xi\nabla_{\mathbf{x}_{\tau}}\mathcal{J}(\hat{\mathbf{P}}).
\label{eq:app_waypoint_guidance}
\end{equation}
We differentiate through the velocity prediction, allowing the correction to adjust the root trajectory together with the object trajectory.
We use $\xi=40$ and $\Delta\tau=1/20$, starting guidance after the first Euler step.

\section{On-Policy Post-Tuning}
\label{app:post_tuning}

\subsection{Velocity-Field Regularization}
\label{app:velocity_regularization}
Let $\mathcal{I}_{\mathrm{leg}}$ contain the $24$ action coordinates of the hips, knees, ankles, and toes.
For a timestep sampled from an on-policy rollout, we use the normalized action $\mathbf{a}_t$ as the target of the flow interpolation in Eq.~\eqref{eq:flow_matching_loss}.
Both the trainable field and the frozen behavior-cloned field are evaluated at the same noisy action, flow time, and conditioning $(\mathbf{C}_t,\mathbf{P}_t,y,g)$.

Writing the trainable and frozen fields' output coordinates as $v_{\theta,j}$ and $v_{\mathrm{BC},j}$, the penalty is
\begin{equation}
\mathcal{L}_{\mathrm{vel}}
=\mathbb{E}_{t,\tau,\mathbf{z}}\!\left[
\frac{1}{24}\sum_{j\in\mathcal{I}_{\mathrm{leg}}}
\left(v_{\theta,j}-v_{\mathrm{BC},j}\right)^2\right].
\label{eq:app_velocity_regularization}
\end{equation}
The expectation averages on-policy timesteps and the Gaussian noise and uniform flow times used by flow matching.

We add $10^{-3}\mathcal{L}_{\mathrm{vel}}$ to the actor loss; the frozen field is evaluated with LoRA disabled and receives no gradients.
Only the leg coordinates are constrained by this loss, preserving their pretrained action patterns while allowing the arms and hands to adapt.

\subsection{Optimization Settings}
LoRA modifies attention query, value, and output projections, feed-forward layers, and the action projection.
The critic has widths $(2048,1024,512)$ and observes the state, text, committed plan, goal, following errors, and completion state.

The critic receives $20$ critic-only rollout iterations before joint updates begin.
Training starts at frame zero with probability $0.3$, or uniformly within the first $70\%$ of a retained rollout otherwise.
We report additional training configs in Table.~\ref{tab:post_tuning}.

\begin{table}[t]
\centering
\small
\caption{\textbf{Additional post-tuning optimization settings.}}
\label{tab:post_tuning}
\begin{tabular}{@{}p{0.62\columnwidth}p{0.32\columnwidth}@{}}
\toprule
Setting & Value \\
\midrule
Parallel environments / rollout steps & $16{,}384/32$ \\
Update epochs per rollout & $4$ \\
Optimizer / critic learning rate & AdamW / $10^{-4}$ \\
Weight decay / gradient-norm clipping & $10^{-4}/1$ \\
Discount / GAE parameter & $0.99/0.95$ \\
PPO clip / value-loss coefficient & $0.2/2$ \\
Entropy coefficient & $0$ \\
Flow-SDE steps / noise level & $5/0.5$ \\
Maximum episode duration & $10$\,s \\
\bottomrule
\end{tabular}
\end{table}

\subsection{Motion Prior}
\label{app:motion_prior}
\paragraph{Training data and representation.}
We train SMP~\citep{mu2025smp} on retargeted OMOMO motion capture, excluding subject 14.
The motion prior's $20$-frame windows contain $156$ character features per frame: root pose and velocities ($15$), local 6D rotations of 21 non-finger joints ($126$), and head, wrist, and ankle positions ($15$).
The window shares the last frame's root heading and horizontal origin, retaining absolute height.
Velocities are obtained by finite differences, and all channels are standardized using training-set statistics.

\paragraph{Prior training and reward.}
The prior predicts the Gaussian noise injected into a motion window, trained with an $\ell_1$ prediction loss and a $50$-step cosine DDPM schedule.
The transformer has three blocks, width $320$, five heads, and feed-forward width $1280$, with frozen CLIP text embeddings supplied through cross-attention.

We use AdamW for $800{,}000$ updates with batch size $512$, learning rate $2\times10^{-4}$, weight decay $0.01$, and EMA decay $0.995$.
Text augmentation follows Stage II.

We compute motion prior rewards following SMP~\citep{mu2025smp}.
For our implementation, we use diffusion steps $\{8,15,22\}$ and estimate the error-normalization statistics from simulated motion windows during critic warmup.
These statistics remain fixed throughout subsequent post-tuning.

\begin{table}[t]
\centering
\small
\caption{\textbf{Post-tuning reward coefficients.}}
\label{tab:rl_weights}
\begin{tabular}{@{}p{0.62\columnwidth}p{0.32\columnwidth}@{}}
\toprule
Term & Weight \\
\midrule
$w_{\mathrm{task}}/w_{\mathrm{motion}}$ & $0.5/0.5$ \\
$b_{\mathrm{goal}}$ & $100$ \\
$\lambda_{\mathrm{obj}},\lambda_{\mathrm{root}}$ & $(0.15\,\mathrm{m})^{-2}$ each \\
Grasp / palm / hold & $0.15/0.35/0.05$ \\
Hand-contact sharpness & $5$ \\
Joint / object acceleration & $2\times10^{-5}$ each \\
Contact force / reference anchor & $10^{-9}/0$ \\
\bottomrule
\end{tabular}
\end{table}

\subsection{Reference-Free Terminations}
\label{app:terminations}

\paragraph{Plan divergence.}
We exponentially smooth the object and root following errors over approximately one second and half a second, respectively, so that brief deviations do not immediately end an episode.
We terminate when the smoothed object error exceeds $0.30\,\mathrm m$ or the root error exceeds $0.25\,\mathrm m$.

\paragraph{Insufficient object progress.}
For each plan, we compare how far the object has moved with how far the plan requested it to move.
We measure both displacements using $d^{\mathrm{obj}}$, taking the largest displacement reached so far so that returning toward the starting pose does not erase earlier progress.
Denoting the requested and achieved displacements by $D_j^{\mathrm{plan}}(t)$ and $D_j^{\mathrm{ach}}(t)$, the achieved fraction is
\begin{equation}
u_j(t)=\min\!\left\{
\frac{D_j^{\mathrm{ach}}(t)}{D_j^{\mathrm{plan}}(t)},1
\right\}.
\label{eq:app_progress}
\end{equation}
We average this fraction across active plans requesting at least $2\,\mathrm{cm}$ of motion and smooth the result over one second, starting from one.
We terminate when the smoothed fraction falls below $0.40$, indicating that the object consistently achieves less than $40\%$ of the requested motion.

\paragraph{Fall and completion.}
A pelvis height below $0.15$\,m triggers a fall termination.
Successful completion requires the object to remain within $0.20$\,m of its goal under $d^{\mathrm{obj}}$ for $15$ steps ($0.5$\,s).

\subsection{Component Ablation}
\label{app:post_tuning_ablation}

\begin{table*}[t]
\centering
\caption{\textbf{Ablation of on-policy post-tuning components on OMOMO.}
All variants are initialized from the same Flow-BC checkpoint and remove only the indicated component.
For the ablation without object-progress termination, this termination criterion is disabled during both post-tuning and evaluation.
Bold indicates the best value among variants using the standard evaluation termination criteria.
}
\label{tab:flow_rl_ablation}
{\setlength{\tabcolsep}{3.5pt}
\begin{adjustbox}{max width=\textwidth}
\begin{tabular}{lcc cccc cccccc}
\toprule
& \multicolumn{2}{c}{\bf Task}
& \multicolumn{4}{c}{\bf Semantic}
& \multicolumn{6}{c}{\bf Motion quality} \\
\cmidrule(lr){2-3}
\cmidrule(lr){4-7}
\cmidrule(lr){8-13}
Method
& SR (\%)\,$\uparrow$
& Time (fr)
& R@1\,$\uparrow$
& R@3\,$\uparrow$
& MM-Dist\,$\downarrow$
& Div.\,$\rightarrow$
& Penet.\,$\downarrow$
& Float\,$\downarrow$
& Skate\,$\downarrow$
& Phys-Err\,$\downarrow$
& Jerk\,$\downarrow$
& NDMS\,$\uparrow$ \\
\midrule

\rowcolor{gray!20}
\multicolumn{13}{c}{\bf Train split (3485 sequences)} \\

\textbf{Flow-BC + FT}
& \textbf{91.33}
& 135.2
& 0.374
& 0.698
& 4.61
& \textbf{15.08}
& 0.02
& \textbf{13.25}
& 7.19
& \textbf{20.46}
& 281.7
& 0.452 \\

w/o trajectory-following costs
& 81.81
& 141.4
& 0.381
& 0.710
& 4.60
& 14.19
& 0.04
& 13.69
& \textbf{6.74}
& 20.46
& 272.2
& 0.446 \\

w/o motion prior
& 87.72
& 137.1
& 0.364
& 0.693
& 4.56
& 14.65
& \textbf{0.00}
& 13.70
& 6.90
& 20.59
& 308.0
& 0.448 \\

w/o completion bonus
& 80.69
& 145.3
& \textbf{0.405}
& \textbf{0.736}
& \textbf{4.19}
& 14.66
& 0.02
& 13.64
& 6.84
& 20.49
& \textbf{252.6}
& 0.456 \\

w/o hand-interaction costs
& 88.69
& 138.2
& 0.390
& 0.722
& 4.35
& 14.80
& 0.01
& 13.71
& 7.15
& 20.81
& 269.2
& \textbf{0.458} \\

w/o object-progress termination
& 84.85
& 247.6
& 0.362
& 0.675
& 4.94
& 14.32
& 0.02
& 12.80
& 5.88
& 18.68
& 228.4
& 0.425 \\

\midrule

\rowcolor{gray!20}
\multicolumn{13}{c}{\bf Held-out split (267 sequences)} \\

\textbf{Flow-BC + FT}
& \textbf{90.26}
& 148.8
& 0.361
& 0.666
& 4.94
& --
& 0.02
& 7.08
& 6.75
& 13.85
& 256.7
& 0.451 \\

w/o trajectory-following costs
& 77.53
& 160.8
& 0.357
& 0.686
& 4.76
& --
& \textbf{0.00}
& 8.07
& \textbf{6.11}
& 14.18
& 247.8
& 0.437 \\

w/o motion prior
& 84.64
& 152.4
& 0.313
& 0.633
& 5.10
& --
& 0.03
& 7.22
& 6.38
& 13.63
& 284.4
& 0.444 \\

w/o completion bonus
& 74.91
& 167.7
& \textbf{0.415}
& \textbf{0.731}
& \textbf{4.22}
& --
& 0.03
& 7.11
& 6.13
& 13.27
& \textbf{226.4}
& 0.448 \\

w/o hand-interaction costs
& 85.77
& 161.3
& 0.371
& 0.719
& 4.49
& --
& \textbf{0.00}
& \textbf{6.72}
& 6.16
& \textbf{12.88}
& 233.5
& \textbf{0.454} \\

w/o object-progress termination
& 81.65
& 307.0
& 0.350
& 0.660
& 4.98
& --
& 0.00
& 6.55
& 5.08
& 11.62
& 200.5
& 0.411 \\

\bottomrule
\end{tabular}
\end{adjustbox}
}
\end{table*}

Table~\ref{tab:flow_rl_ablation} evaluates five variants
initialized from the same Flow-BC checkpoint.
\emph{w/o trajectory-following costs} sets
$\lambda_{\mathrm{obj}}=\lambda_{\mathrm{root}}=0$
while retaining the plan-divergence terminations.
\emph{w/o motion prior} sets $w_{\mathrm{motion}}=0$
and increases $w_{\mathrm{task}}$ from $0.5$ to $1$
to preserve the overall weighting of the dense reward.
\emph{w/o completion bonus} sets $b_{\mathrm{goal}}=0$
and enables value bootstrapping at successful termination.
\emph{w/o hand-interaction costs} removes the physical
hand--object contact, geometry-conditioned grasp,
palm-alignment, and hold-stability costs from
$E_t^{\mathrm{int}}$, while retaining the energy costs.
\emph{w/o object-progress termination} disables this
criterion during both post-tuning and evaluation,
while retaining the plan-divergence and fall terminations.
The complete objective achieves the highest success on both the training
and held-out splits, at $91.33\%$ and $90.26\%$, respectively.
Removing either the trajectory-following costs or the completion bonus causes the
largest reductions, showing that dense short-horizon guidance and sparse
completion supervision provide complementary task signals.
Interestingly, weakening these objectives improves several mocap-based
semantic and motion-quality metrics.
This reflects a trade-off between reference-distribution similarity and
physical task completion: corrective reaching, regrasping, and terminal
placement may depart from common mocap patterns, whereas smoother motions
that avoid such corrections can score better without successfully
manipulating the object.

Removing the motion prior degrades retrieval-based semantic alignment and temporal motion quality, as reflected by lower R-Precision and NDMS and higher Jerk. 
Although the motion-prior reward is introduced to encourage natural humanoid motion, the variant without it also exhibits lower task success.
Inspection of policy rollouts suggests that the text-conditioned motion prior provides dense guidance during the pre-contact phase, when the unactuated object has not yet moved and the object-plan-following term offers little signal for approaching the object and initiating contact. 
Without this guidance, the policy sometimes remains near the object without initiating the interaction. 
These observations suggest that the motion prior serves not only as a style regularizer, but also as an implicit behavioral prior for object approach and contact acquisition. 

Removing the hand-interaction costs also reduces task success, despite improving several generic motion-quality metrics. 
This contrast shows that smooth, mocap-like motion alone does not guarantee a functional and stable grasp, highlighting the importance of explicitly optimizing interaction quality.

Finally, disabling object-progress termination during both post-tuning
and evaluation roughly doubles rollout duration but still reduces
success.
Longer execution therefore does not recover stalled interactions; the
progress criterion complements absolute plan-following errors by detecting
insufficient object motion under small incremental plans.

\section{Evaluation Details}
\label{app:evaluation}

\subsection{Text--Motion Evaluator}
\label{app:evaluator}
\paragraph{Data and representation.}
Evaluator training uses retargeted motion capture before filtering by teacher success.
We exclude subject~14 and split the remaining clips into 4,337 training and 228 validation examples.
Each clip's original caption and eight paraphrases remain in the same split; a caption is sampled randomly during training.

The essential-motion representation~\citep{meng2025rethinking} contains four root channels---yaw velocity, horizontal velocity, and height---and the root-relative positions of 21 body joints, giving 67 dimensions.
Positions are expressed in the humanoid heading frame.
We compute normalization statistics on training clips.

\paragraph{Training.}
Following Guo et al.~\citep{guo2022generating}, training has two stages.
First, a temporal convolutional autoencoder reconstructs 40-frame windows with an $\ell_1$ reconstruction loss and penalties on latent magnitude and adjacent-frame latent differences.
The autoencoder's movement encoder reduces temporal resolution by a factor of four.

In the second stage, we freeze the movement encoder and train text and motion BiGRUs to bring matched pairs together and separate mismatched pairs with a contrastive margin of 10.
The text and motion branches produce 512-dimensional embeddings; text inputs use 300-dimensional GloVe vectors and part-of-speech features.

Table~\ref{tab:app_evaluator} lists the settings.
Both stages use Adam, and the checkpoint with the lowest validation loss is selected for each stage.
The resulting evaluator remains fixed across all comparisons.

\begin{table}[t]
\centering
\caption{\textbf{Text--motion evaluator training settings.}}
\label{tab:app_evaluator}
\small
\begin{tabular}{p{0.43\columnwidth}p{0.19\columnwidth}p{0.19\columnwidth}}
\hline
Setting & Autoencoder & Matching \\
\hline
Batch size & 128 & 32 \\
Learning rate & $10^{-4}$ & $10^{-4}$ \\
Maximum epochs & 270 & 300 \\
Motion window / maximum length & 40 frames & 300 frames \\
Text / motion BiGRU hidden width & --- & $512/1024$ \\
Latent magnitude / smoothness weights & $10^{-3}$ / $10^{-3}$ & --- \\
Gradient clipping & --- & 0.5 \\
\hline
\end{tabular}
\end{table}

\paragraph{Scoring and rollout eligibility.}
We use 20 repetitions, retrieval batches of 32, and 300 sampled pairs for Diversity.
The repetitions resample caption and motion windows and randomize retrieval groups.
Rollouts shorter than 40 feature frames are excluded, and incomplete retrieval batches are dropped, so semantic metrics can use different subsets for different methods.

\subsection{Kinematic Generation-and-Tracking Baselines}
\label{app:kinematic_baselines}
\paragraph{Motion representation.}
The generated representation contains 9-dimensional root and object poses and 153 joint coordinates, totaling 171 dimensions.
Forward kinematics reconstructs the body transforms needed by the tracker.
Contact labels are inferred with a $0.05\,\mathrm{m}$ distance threshold, using the same rule on training motion and the simulated histories supplied to Replan. kin. planner.

\paragraph{Training.}
We train the kinematic generator on retargeted motion capture excluding subject~14 for one million flow-matching updates, using AdamW, learning rate $10^{-4}$, batch size 512, and EMA decay 0.9999.
The trackers for both baselines initialize from the same unified teacher; their fine-tuned variants use the common completion-bonus weight in Table~\ref{tab:rl_weights}.
The generator and tracker actor together contain 126.6 million parameters, compared with 127.0 million for our planner and action generator.

\subsection{Novel Object Shapes}
\label{app:novel_shapes}
\begin{figure*}[t]
    \centering
    \includegraphics[width=0.85\textwidth]{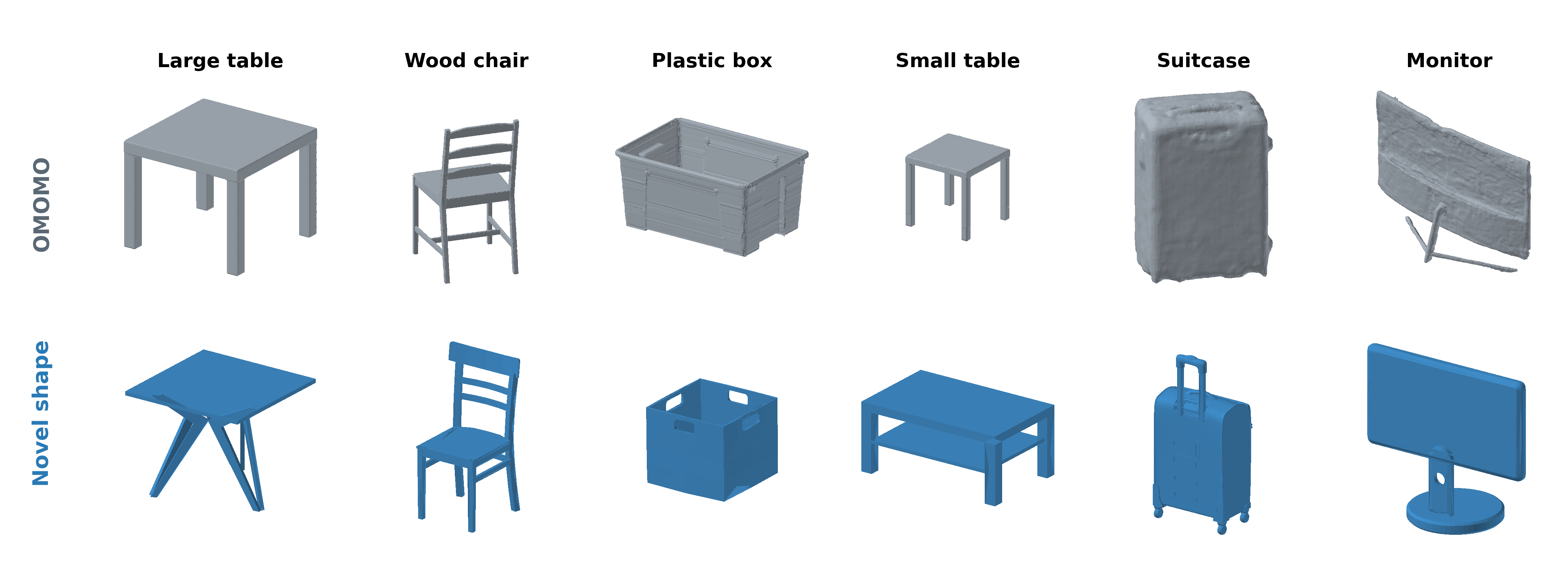}
    \caption{\textbf{Original and novel object geometry.}
    Original OMOMO meshes (top) and the exact Objaverse replacements used in selected successful videos (bottom).
    Each column uses the same view and scale; scale varies across categories.
    }
    \label{fig:app_shape_comparison}
\end{figure*}

Fig.~\ref{fig:app_shape_comparison} shows some examples of the novel object shape meshes compared with the original dataset's meshes. 
For each replacement mesh, we recompute both the collider and the surface-distance representation, so geometric observations use the new shape.

\section{Human Study}
\label{app:human_study}
\begin{figure*}[t]
    \centering
    \begin{minipage}[t]{0.49\textwidth}
        \centering
        \includegraphics[width=\linewidth]{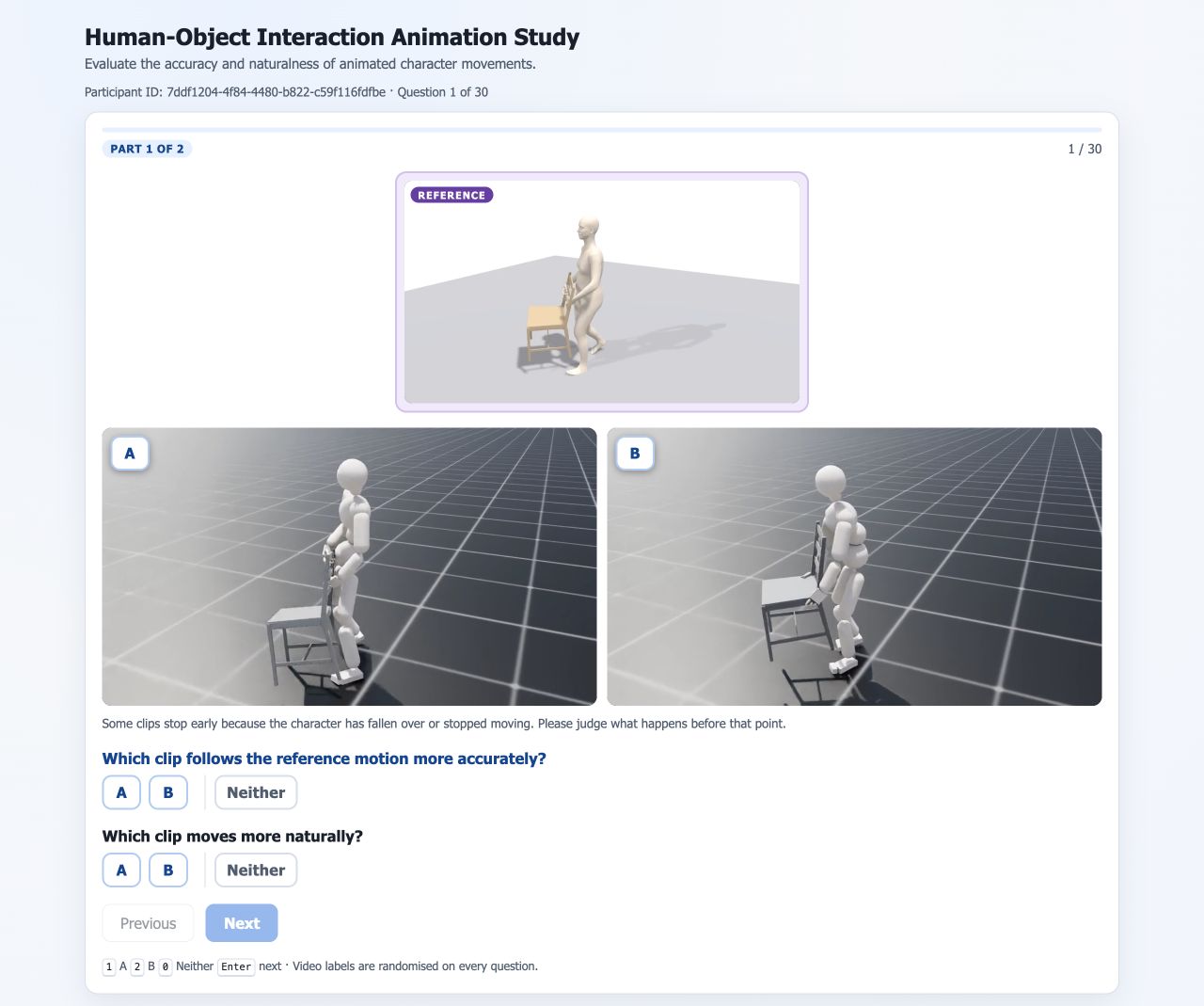}
        \par\smallskip
        {\small (a) Tracking study}
    \end{minipage}
    \hfill
    \begin{minipage}[t]{0.49\textwidth}
        \centering
        \includegraphics[width=\linewidth]{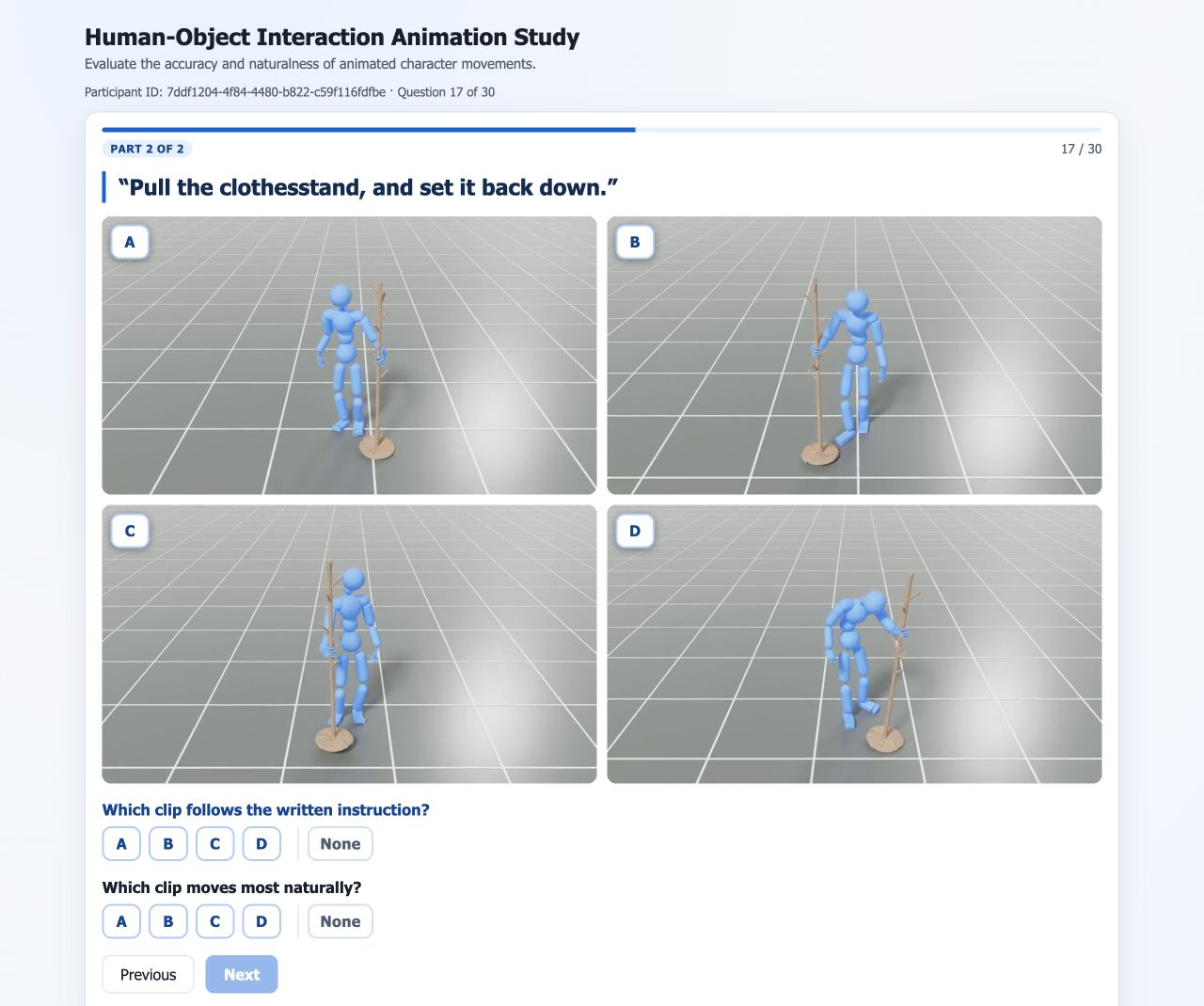}
        \par\smallskip
        {\small (b) Generation study}
    \end{minipage}
    \caption{\textbf{Human-study interfaces.}
    (a) Participants compare tracking fidelity and naturalness given a motion-capture reference.
    (b) Participants compare instruction adherence and naturalness among four anonymous candidates given a text instruction.
    }
    \label{fig:app_human_study}
\end{figure*}

\paragraph{Protocol and questions.}
The study included 14 participants, each of whom completed both blocks: 10 tracking trials followed by 20 generation trials. 
In total, this yields 140 tracking trials and 266 generation trials after excluding attention-check trials.
Trial order is randomized within each block, and the mapping from methods to anonymous candidate letters is randomized independently for every trial.
Each trial asks for two separate selections, one for fidelity and one for naturalness.
For each question, participants choose a single candidate or \emph{Neither}/\emph{None}.
Fig.~\ref{fig:app_human_study} shows the conditioning material, candidates, and exact questions.

\end{document}